*Review Paper*

Moein Razavi[a,b,f], Samira Ziyadidegan[a], Ahmadreza Mahmoudzadeh[e,f], Saber Kazeminasab[c], Elaheh Baharlouei[d], Vahid Janfaza[b], Reza Jahromi[a,b], Farzan Sasangohar[a,*]

[a] Department of Industrial and Systems Engineering, Texas A&M University
[b] Department of Computer Science and Engineering, Texas A&M University
[c] Harvard Medical School, Harvard University
[d] Department of Computer Science, University of Houston
[e] Engineering Academic and Student Affairs, College of Engineering, Texas A&M University
[f] Ford Motor Company, Global Data Insight & Analytics

Corresponding author. 3131 TAMU, College Station, TX, 77843, USA. E-mail address: sasangohar@tamu.edu

# Machine Learning, Deep Learning and Data Preprocessing Techniques for Detection, Prediction, and Monitoring of Stress and Stress-related Mental Disorders: A Scoping Review

## *Abstract*

**Background**: Mental stress and its consequent mental disorders (MDs) constitute a significant public health issue. With the advent of machine learning (ML), there's potential to harness computational techniques for better understanding and addressing mental stress and MDs. This comprehensive review seeks to elucidate the current ML methodologies employed in this domain to pave the way for enhanced detection, prediction, and analysis of mental stress and its subsequent mental disorders.

**Objective**: This review aims to investigate the scope of Machine Learning (ML) methodologies employed in the detection, prediction, and analysis of mental stress and its consequent mental disorders (MDs).

**Methods**: Utilizing a rigorous scoping review process with Preferred Reporting Items for Systematic Reviews and Meta-Analyses extention for Scoping Reviews (PRISMA-ScR) guidelines, this investigation delves into the latest ML algorithms, preprocessing techniques, and data types employed in the context of stress and stress-related MDs.

**Results and Discussion**: Total of 98 peer-reviewed publication were examined for this review. The findings highlight that Support Vector Machine (SVM), Neural Network (NN), and Random Forest (RF) models consistently exhibit superior accuracy and robustness among all machine learning algorithms examined. Physiological parameters such as heart rate measurements and skin response are prevalently used as stress predictors due to their rich explanatory information concerning stress and stress-related MDs, as well as the relative ease of data acquisition. The application of dimensionality reduction techniques, including mappings, feature selection, filtering, and noise reduction, is frequently observed as a crucial step preceding the training of ML algorithms.

**Conclusion**: The synthesis of this review identifies significant research gaps and outlines future directions for the field. These encompass areas such as model interpretability, model personalization, the incorporation of naturalistic settings, and real-time processing capabilities for the detection and prediction of stress and stress-related MDs.

**Keywords:** Machine Learning; Deep Learning; Data Preprocessing; Stress Detection; Stress Prediction; Stress Monitoring; Mental Disorders

## *Introduction*

Mental health has become a public health concern. According to Institute of Health Metrics and Evaluation (IHME), in 2019, about 53 million people in the United States and about one in eight individuals worldwide (about 1 billion people) suffer from at least one mental health disorder (MD) [1]. MD is defined as an impairment in a person's cognition, emotional control, or behavior pattern, which has clinical significance and is often linked to distress or functional impairment [2]. MDs severely limit people's daily functioning and can be fatal [3], [4]. In 2019, mental health (MH) problems accounted for 6.6% of all disability-adjusted life years in the US, making it the fifth most significant cause of disability overall [5], [6].

Some of the more prevalent MDs are anxiety disorders, depression or mood disorders, bipolar disorders, psychotic disorders (including schizophrenia), eating disorders, social disorders and disruptive behavior and addictive behaviors [2]. In 2019, anxiety and depression have been the most prevalent forms of MDs (301 and 280 million people affected worldwide, respectively). Anxiety disorder encompasses emotions of concern, anxiety, excessive fear, or associated behavioral problems that are severe enough to affect everyday activities [2]. Symptoms include an unproportionate level of stress compared to the significance of the triggering event, difficulty in putting worries out of one's mind, and nervousness [7], [8]. Generalized anxiety disorder, panic attacks, social anxiety disorder, and post-traumatic stress disorder are all examples of different types of anxiety disorders [2], [9]. Depression is characterized by a long-lasting sadness and a lack of desire to be active. One of the main symptoms of depression is the inability to enjoy or find pleasure in most of one's daily activities as well as felling sadness, anger, or emptiness [2], [10]. A depressive episode typically lasts for at least two weeks. Additionally, a loss of self-worth, feelings of hopelessness for the future and suicidal thoughts are indicators and symptoms of depression. People who are depressed are more prone to commit suicide [2], [10], [11].

Stress is categorized into distress, which typically has chronic negative effects on health, and eustress, which is short-term and positively influences motivation and development [12]. Throughout this paper, the term stress is specifically used to denote distress, rather than eustress. Mental stress has shown to significantly contribute to developing and worsening anxiety and depression disorders [13], [14], [15]. Mental stress is the body's natural response to various events in which a person feels that the demands of their external environment exceed their psychological and physiological resources for dealing with those demands [16]. Mental stress leads to an asynchrony between the sympathetic and parasympathetic nervous systems (SNS and PNS) which are the main divisions of the autonomic nervous system (ANS) [17] and serve an important role in regulating vital biological activities [18], [19]. The sympathetic nervous system is an integrative system that responds to potentially dangerous circumstances. Activation of the sympathetic nervous system is part of the system responsible for controlling ‘fight-or-flight’ responses. The parasympathetic nervous system is responsible for the body's "rest and digest" processes.

Given the import role and impact of stress in MDs, previous research has investigated various qualitative and quantitative methods to measure and monitor stress to inform effective stress mitigation approaches. While majority of stress literature relies on self-reported measures, recent literature has used physiological variables such as heart rate, heart rate variability [20], [21], [22], [23], [24], and behavioral data (e.g., speech, movement, facial expressions) [25] to understand changes to SNS and PNS associated with stress. The recent advances in sensor and mobile health technologies has resulted in the emergence of “big data” related to mental health as well as advanced bioinformatics methods, tools, or techniques to use such data for modeling or inference. One such tool that has recently emerged as a robust, rapid, objective, reliable, and cost-efficient technique for studying chronic illnesses and MDs is Machine Learning (ML). ML uses advanced statistical and probabilistic techniques to construct systems that can automatically learn from data. Several characteristics of ML makes it suitable for applications in MH monitoring including significant pattern recognition and forecasting capabilities [26], capacity to extract crucial information from various data resources and opportunity to create personalized experiences [26], and ability to analyze large amounts of data in a short time [27]. As such, ML has gained popularity and has been applied to MH data to enable detection, monitoring, and treatment [28]. The objective of this research is to review the literature to summarize and synthesize the application of ML in the detection, monitoring, or prediction of stress and stress-related MDs, in particular anxiety and depression. This paper documents methods-specific findings such as data types, preprocessing methods, and different algorithms used as well as type and characteristics of studies that used ML. Traditional statistical methods, such as linear regression, logistic regression, t-tests, and ANOVA [29], have been widely employed in the past to detect and analyze stress and stress-related MDs. These methods have proven useful in specific contexts, such as comparing means of different groups, or modeling linear relationships between variables. As demonstrated by [22], [23], [24], [25] and [26], these methods have provided valuable insights in situations where the data is relatively simple and adheres to the underlying assumptions of the statistical techniques. However, when faced with complex, high-dimensional mental health data, which has become increasingly available thanks to advancements in technology and data collection techniques, these traditional statistical methods might not be sufficient. The limitations of these methods stem from their inherent simplicity and the assumptions they rely on, which might not hold true in the context of MH data. For example, linear and logistic regression assume linear relationships between variables, while t-tests and ANOVA require specific assumptions about the data distribution. These assumptions may not be applicable in the case of intricate and heterogeneous MH data, potentially leading to inaccurate or incomplete conclusions.

Advanced data analytics methods, such as machine learning (ML), offer a more powerful and flexible alternative to traditional statistical methods. ML algorithms, with their significant pattern recognition and forecasting capabilities [26], are capable of capturing complex, nonlinear relationships between variables and can adapt to various data distributions. These capabilities enable ML techniques to provide more accurate and insightful predictions, classifications, and associations in the context of MH data [34]. Additionally, ML algorithms can handle large-scale, high-dimensional data more efficiently than traditional methods, allowing researchers to analyze vast amounts of information from diverse sources, such as electronic health records, wearable devices, and online platforms [27]. This capacity for handling big data is crucial for understanding the multifaceted nature of mental health disorders and developing tailored interventions. ML techniques also offer the advantage of automation and adaptability, allowing them to continuously learn and improve as new data becomes available [26]. This iterative learning process enables the development of more sophisticated and accurate models for detecting, monitoring, and predicting stress and stress-related MDs over time.

While traditional statistical methods have contributed significantly to our understanding of stress and stress related MDs in specific contexts, the growing complexity and volume of MH data necessitate the adoption of advanced data analytics methods like ML. By leveraging the power of ML, researchers can gain deeper insights into the underlying patterns and relationships between stress and MDs [34], ultimately leading to the development of more effective stress mitigation approaches and improved care for individuals suffering from anxiety, depression, and other MDs.

Acknowledging the substantial contributions of traditional statistical methods, it becomes evident that the escalating complexity and scale of mental health data demands the adoption of more sophisticated approaches such as ML. This advancement stands not as a replacement but as an essential evolution in the analytical toolbox available to researchers. As this paper delves into the myriad ways ML has been applied to mental health, particularly in the realms of stress, anxiety, and depression, it seeks to consolidate the current knowledge on the subject. By examining the types of data, preprocessing methods, and the algorithms used in existing studies, this review aspires to offer a detailed synthesis of the field. It aims to provide a clearer understanding of ML's effectiveness in the detection, monitoring, and prediction of mental health disorders, setting a foundation for future research and the enhancement of therapeutic strategies for those impacted by these conditions.

## *Methods*

### Protocol and Registration

This scoping review adhered to the Preferred Reporting Items for Systematic reviews and Meta-Analyses extension for Scoping Reviews (PRISMA-ScR) guidelines [35]. No formal review protocol was registered due to the exploratory nature of this study, which aimed to map out existing research rather than address a prespecified hypothesis. This approach aligns with the methodological flexibility often required in emergent areas of research.

### Eligibility Criteria

We included studies published in English from 2017 to 2022 that utilized machine learning (ML) techniques to evaluate mental health disorders, specifically focusing on stress and stress-related conditions. Studies were excluded if they did not use ML as the primary analysis method or if they were published in languages other than English.

### Information Sources

The literature search involved databases such as EI Engineering Village, Web of Science, ACM Digital Library, and IEEE Xplore. Additional sources were identified through contact with experts and review of references in relevant articles.

### Search Strategy

A comprehensive search was conducted using a combination of keywords related to ML and mental health disorders (Table 1). The search strategy was designed to capture a broad spectrum of ML applications within this field. The full search list from all databases is available in the Multimedia Appendix.

**Table 1**. Keywords and search strategy for articles since 2017 (last 5 years)

| First keyword | | Second keyword | | Third keyword |
|---|---|---|---|---|
| predict OR detect | AND | mental health OR mental disorder OR depression OR anxiety OR stress | AND | machine learning OR deep learning OR data mining OR pattern classification OR artificial intelligence OR neural networks |

### Study Selection, Inclusion, and Exclusion Criteria

Articles that did not fully use ML for stress or stress related MDs evaluations were excluded from the research. Studies published in languages other than English were also excluded. The initial search yielded 1241 results. After duplicate articles were deleted and eligibility was confirmed using Rayyan QCRI [36], 1204 articles remained. After applying the exclusion criteria, 98 papers were selected for full review (Figure 1).

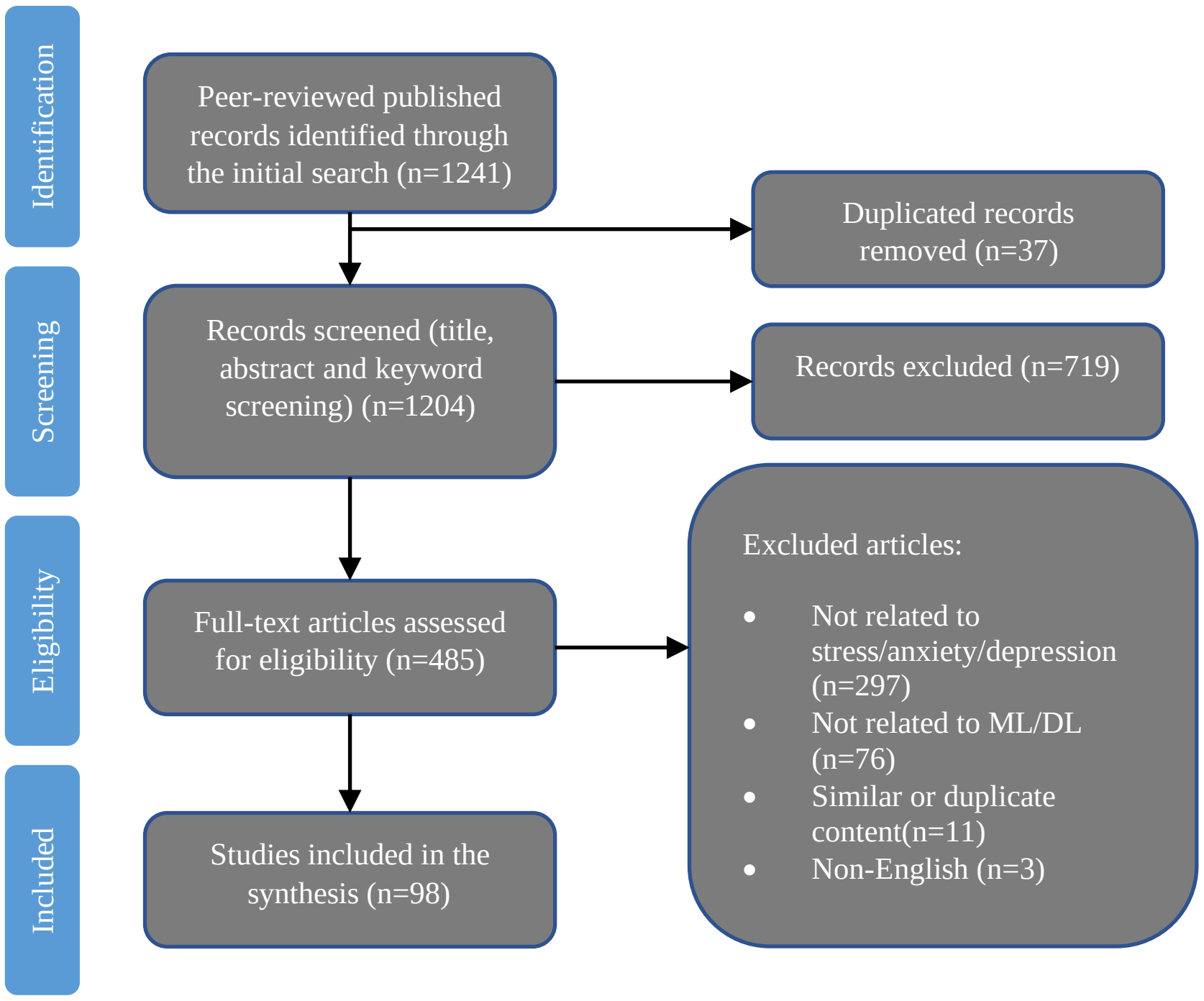

**Figure 1.** Preferred items for scoping literature review and meta-analysis flowchart [35]

### Data Charting Process

Data charting was conducted by two reviewers independently using a standardized form, which had been pretested on a subset of included studies. Discrepancies were resolved through discussion or consultation with a third reviewer. Study authors were contacted for clarification or additional data where necessary.

### Data Items

Data extracted included publication year, study design, population characteristics, ML techniques used, outcomes measured, and key findings. Other variables sought included data preprocessing methods and performance metrics of the ML models. Simplifying assumptions, such as considering different ML algorithms within the same family as a single technique, were made to facilitate synthesis.

### Synthesis of Results

Data were synthesized descriptively, grouping findings by ML techniques, data type and preprocessing techniques. Where possible, quantitative performance metrics were extracted or derived. Results were analyzed in the context of the overall study designs and populations to highlight trends and identify gaps in the current research landscape. No formal critical appraisal or quantitative meta-analysis was conducted due to the diversity of the included studies and the scoping nature of this review.

## *Results and Discussion*

In this section types of data, preprosessing techniques, and ML techniques used on the data in the literature have been reviewed, and compared with the existing literature.

## Types of Data

Various data types were used in the studies that used ML algorithms for stress and stress-related MDs. Studies used questionnaire (n=31), heart rate variability (HRV, n=25), skin response (e.g., skin temperature, skin conductance, etc., n=24), photoplethysmogram (PPG, n=21), electrocardiogram (ECG, n=19), heart rate (HR, n=17), electroencephalogram (EEG, n=9), acceleration/body movement (n=8), text data (n=7), respiratory signals (n=7), electromyogram (EMG, n=3), eye-tracking (n=3), speech signals (n=3) and others (n=4) including audio signals (n=2), blood pressure (BP, n=1), and hormones (n=1). Figure 2 shows the distribution of the type of data used for stress detection using ML techniques.

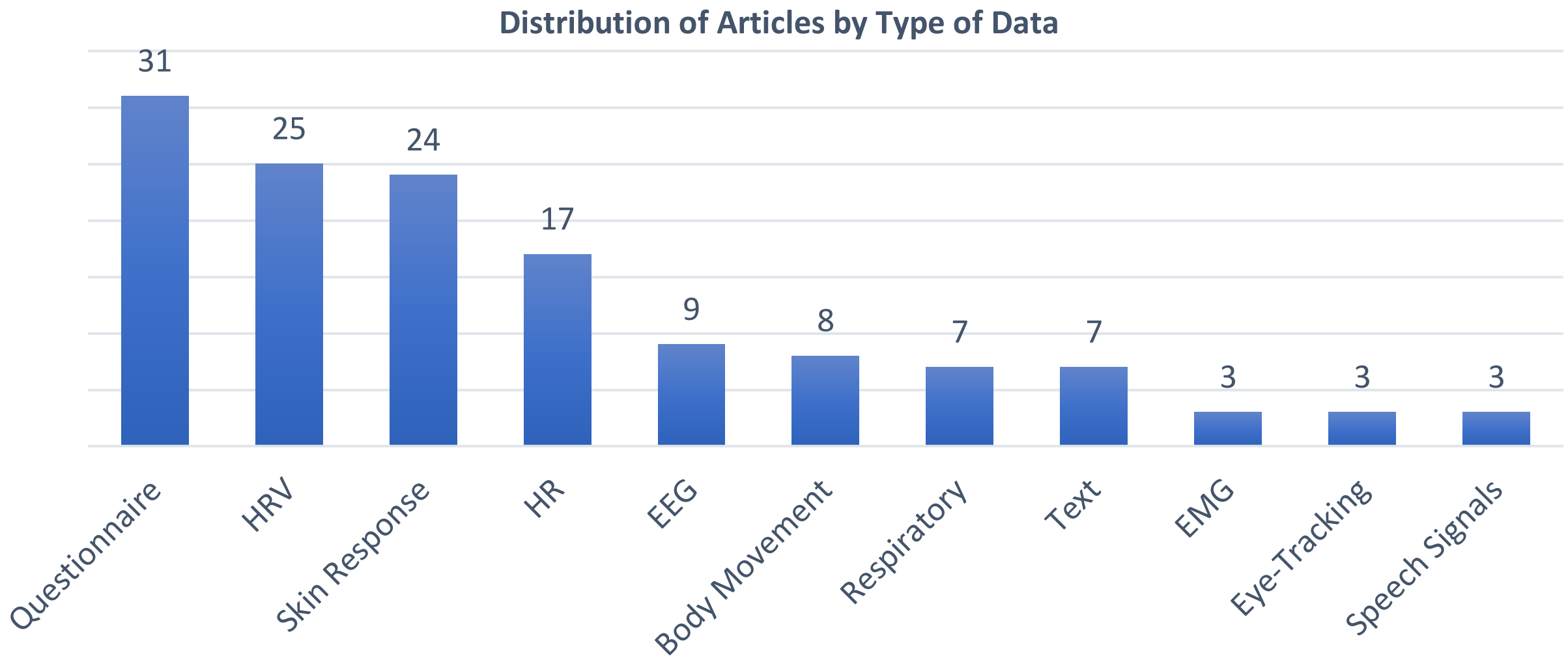

**Figure 2.** Number of articles by types of data

### *Heart Measures*

Heart metrics are primarily utilized for stress detection and are typically gathered through two main methods: ECG (Electrocardiography) and PPG (Photoplethysmography). ECG is a non-invasive diagnostic test that records the heart's electrical activity, while PPG is a non-invasive optical technique that detects changes in blood volume within the tissue's microvascular bed. By employing these methods, it is possible to measure various heart-related parameters, including heart rate (HR), as well as time and frequency domain features of heart rate variability (HRV), and blood pressure (BP).

- *Heart Rate Variability (HRV) (n=25)*: Heart Rate Variability (HRV) has been used to assess mental health issues, such as stress, anxiety, and depression, due to its rich time and frequency domain features [37]. The Blood Volume Pulse (BVP) signal is another effective method for capturing HRV features, as it represents the heart's beat-to-beat volume changes. From the BVP signal, time domain measures like the Root Mean Square of Successive RR Interval Differences (RMSSD), Standard Deviation of NN intervals (SDNN), and Standard Deviation of RR intervals (SDRR) can be derived. Additionally, the frequency domain aspects of HRV, including Total Power (TP, frequencies below 0.4 Hz), Low Frequencies (LF, ranging from 0.04 Hz to 0.15 Hz), and High Frequencies (HF, between 0.15 Hz and 0.4 Hz), reflect the autonomic nervous system's dynamics during beat-to-beat measurements of the heart rate (Figure 3) [38], [39]. These HRV measures, both in the time and frequency domains, provide a nuanced view of the physiological underpinnings associated with various mental health conditions.

- *Heart Rate (HR) (n=17)*: One of the most important indicators of stress is an abrupt increase in HR. Among the physiological signals, HR is among the top measures that explains stress in ML models and it has been used in different studies with almost all ML algorithms [40], [41], [42].

- *Blood Pressure* (BP) *(n=1)*: BP can be obtained by pulse transit time (PTT) or by pressure cuffs [43]. Stressful conditions create an influx of hormones that increase HR and constrict blood vessels leading to a temporary BP elevation [44]. In most cases, BP recovers to its pre-stress level after the stress response diminishes [45]. Schultebraucks et al. used systolic BP as one of the measures in predicting one’s level of susceptibility to Post-Traumatic Stress Disorder (PTSD) [46].

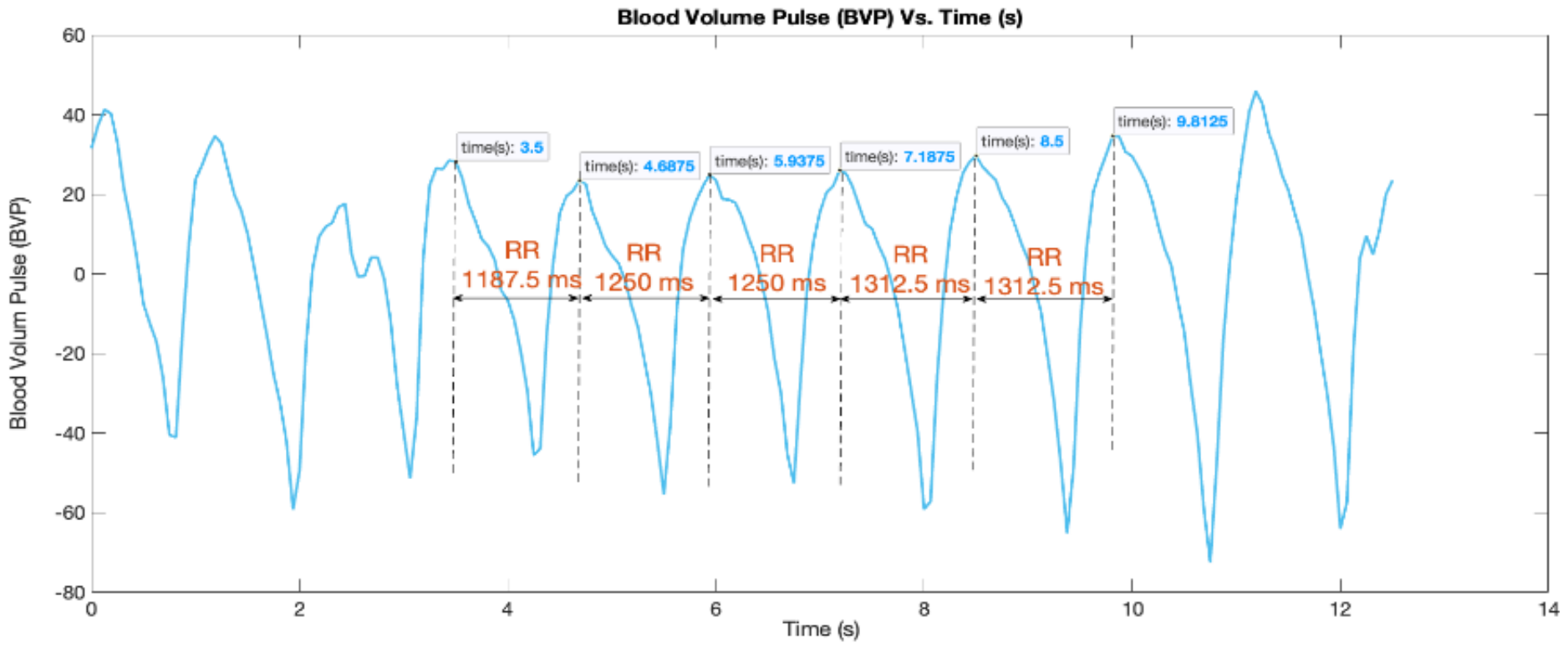

(a)

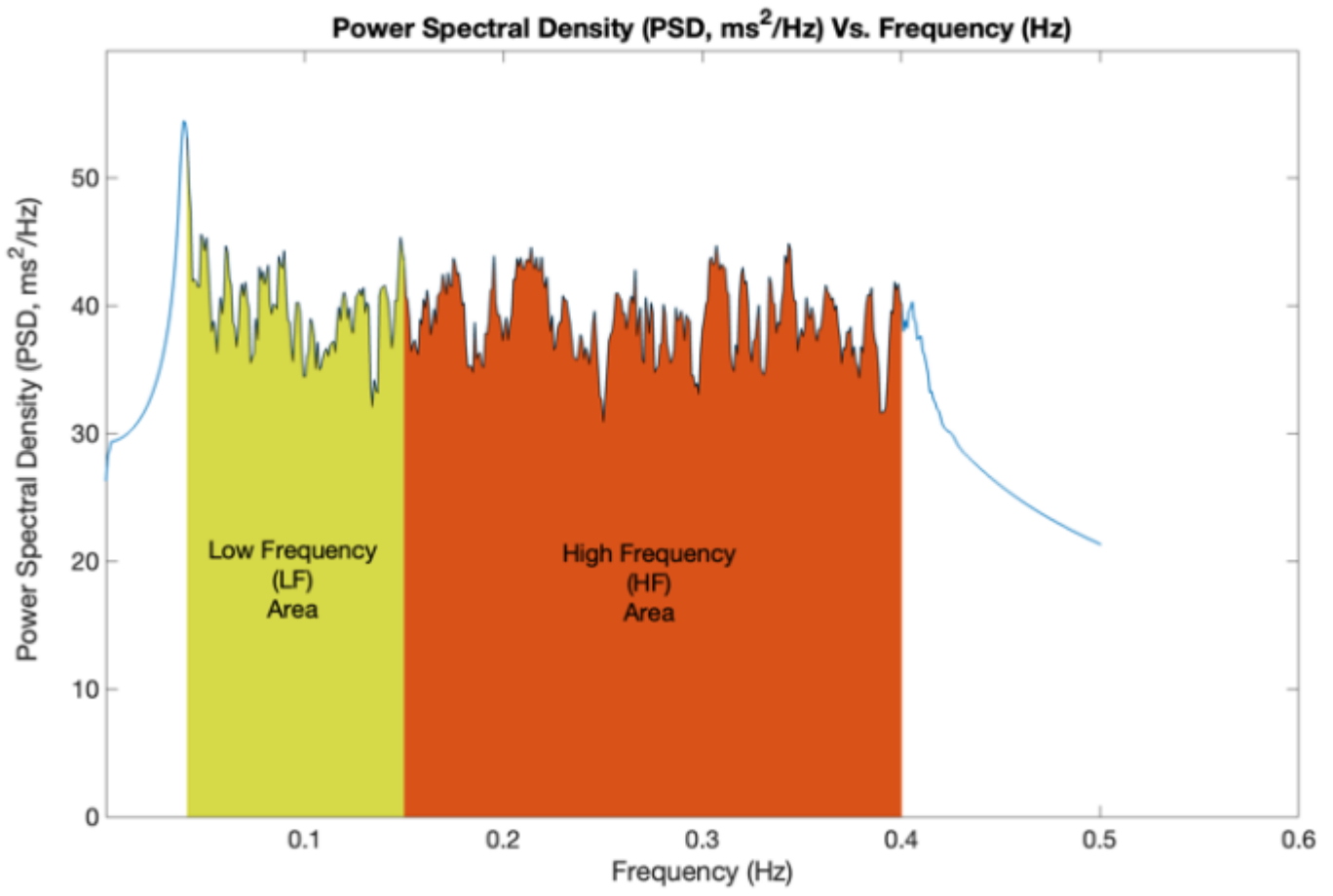

(b)

**Figure 3.** (a) Depiction of heart's beat-to-beat measurements using Blood Volume Pulse (BVP) signal (b) Power Spectral Density (PSD) of RR intervals (the signal is bandpass filtered with cut-off frequencies of 0.04 Hz and 0.4 Hz)

*Electroencephalogram (EEG) (n=9)*: EEG detects brain electrical activity. Compared to other brain mapping techniques, for stress detection, it is more practical due to several factors including affordability, non-invasiveness, non-intrusiveness and most importantly its high temporal resolution *[47]*. The high temporal resolution of EEG makes it appropriate for real-time stress detection, as well as DL approaches which require large dataset for training [47], [48], [49], [50], [51].
The most commonly used EEG features for detection of stress are power of different frequency bands, Alpha (8-13 Hz), Beta (12.5-30 Hz), Theta (4-7.5 Hz), Gamma (30-40 Hz), average and standard deviation of a specific time window of EEG signal, and time-frequency features obtained by Discrete Wavelet Transform (DWT) algorithm [51], [52], [53]. It has also been shown that statistical features of EEG signal such as Kurtosis and Entropy are useful features in stress prediction using ML algorithms [50]. Moreover, Power Spectral Density (PSD), correlation (C), divisional asymmetry (DASM), rational asymmetry (RASM), and power spectrum (PS) are other EEG features that have been used in different studies for stress detection [54].

Since EEG signals are collected from the scalp, they include excessive noise and so they have high uncertainty. Therefore, signal processing and feature selection/extraction is a very important step while dealing with EEG data. Several well-developed methods are available for treating the EEG data. Among them, using latent space derived from auto-encoders and signal reconstruction techniques such as Artifact Subspace Reconstruction (ASR) are well-known methods that can be applied on EEG data to significantly reduce the artifacts [49]. These methods are also fast enough that can make online detection feasible.

Amygdala and hippocampus are the parts of the brain that have the major responsibility for human reactions to stress [55]. Brain activity caused by stress in those regions would affect the prefrontal cortex. Studies collecting data from prefrontal cortex have also verified that EEG data from this brain region can be used for stress detection [56]. EEG can be collected from the prefrontal cortex using off-the-shelf EEG recording products such as MUSE and Neurosky Mindwave [50], [53], [54], [56].

*Eye Tracking (n=3)*: Eye-tracking features can be indicators of stress. For example, to diagnose the level of stress, the changes in the striations of muscle material in the iris as a response to stress can be used as features for ML algorithms. In other words, pupil diameter, which would be controlled by iris sphincter muscles can be used as a feature [57]. Other eye-tracking features that have been for stress detection are visual fixations, saccade movements, pupil size, micro saccades and number of eye-blinks in specific time window during a certain task [58], [59], [60].

*Skin Response (n=24)*: A skin response can be defined as a stimulus-regulated electrodermal response and is typically measured using electrodes placed on the fingertips or hands. Skin response is usually associated with increase in sympathetic activity upon inducing stress events [61]. The skin becomes a better conductor of electricity when it is stimulated either externally or internally by physiologically stimulating factors, including stressful conditions *[62]*.

*Respiratory Signals (n=7)*: Mental stress can affect different respiratory cycle phases and breathing patterns [63], [64]. For example, It is discovered that stress had no impact on overall breath duration (respiration rate), but that exhalation periods were longer and pause periods were shorter in the stress experiment compared to the neutral condition [65].
Based on the findings of several studies, it can be concluded that respiratory signal is one of the top contributing factors in explanation of stress in ML models. The most common time domain respiratory signal features that are extracted for stress detection are: Root Mean Square (RMS), Interquartile range (IQR), Mean of squared Differences between Adjacent elements (MDA) of breathing rate and blood oxygenation levels. The most commonly used frequency domain features of the respiratory signal are the power of low frequencies (LF, under 2 Hz), the power of high frequencies (HF, above 2 Hz) and the ratio of power of low frequencies over the power of high frequencies (LF/HF) [42], [46], [47], [66], [67], [68].

*Electromyogram (EMG) (n=3)*: EMG detects the electrical activity of muscles at rest, during a modest contraction, and during a strong contraction [69]. Similar to acceleration data, several studies have shown that, using EMG data can help increasing the performance of ML models trained on ECG data. The action potential intrigued in the EMG during stress can reduce the variance for decision making of classification models that use ECG [42], [70], [71].

*Hormones (n=1)*: It has been shown that stress can alter the levels of glucocorticoids, catecholamines, growth hormones, and prolactin in the bloodstream. Therefore in ML models, level of hormones such as cortisol, dehydroepiandrosterone sulfate (DHEAS), thyroid-stimulating hormone (TSH), free triiodothyronine (FT3), and free thyroxine (FT4) can be used as predictors for detection of stress-related disorders [46].

*Acceleration/Body Movement (n=8)*: Mental Stress may cause a broad variety of behavioral/body movement symptoms such as shaking hands and feet which can be measured by the acceleration data [72]. Moreover, research has shown that people with a greater stress score had less variance in their activity level and body movements [73], [74], [75]. For example, In the elderly, stressful life events can be related to a reduced rate of regular physical exercise [76]. Time and frequency features such as mean absolute deviation from mean (MAD), total power of acceleration, standard deviation, mean norm of acceleration, absolute integral, peak frequency of each axis are the features of hand/body acceleration used for stress detection [41], [77], [78]. One practical characteristic of motion/acceleration data would be the fact that it can be used to identify" sources of noise in other signals . For example, motion data can help distinguishing stress from physical activity (e.g., exercise) when other physiological measures such as ECG have uncertainty in prediction [79], [80].

*Audio and Speech Signals*

- *Speech Signals (n=3)*: Using speech signals, it is feasible to diagnose and assess neurological and MDs [81]. Moreover, studies have shown that, like body acceleration and EMG, features of speech signal can make stress predictions of heart measurements more robust. The best explanatory parameters of speech signal are frequency domain parameters (e.g., PSD, strongest frequency from FFT transform) and time-frequency features such as Mel-Frequency Cepstral Coefficient (MFCC) [40], [82], [83]. Since time-frequency measures are 2-dimentional measurements with high number of samples, they make this signal suitable for using in convolutional neural network models (CNNs) of stress and depression detection [84].

- *Audio Signals (n=2)*: For lab based studies, audio signals (e.g. beep sounds) can be used for stimulating stress events in participants [85], [86].

*Text data (n=7)*: Social media content is frequently subjected to reviews, opinions, and influence, as well as sentiment analysis. Natural language processing methods may be used to evaluate social networking posts and comments for mood and emotion to detect whether a user is stressed [87], [88], [89], [90], [91], [91], [92], [93].

*Questionnaire (n=31)*: There are different questionnaires that are used for diagnosis of stress and different MDs including anxiety and depression. The scores from different items on these questionnaires can be used as dependent/independent variables in ML studies. The questionnaires mentioned here were selected based on their prevalence in the literature as well as their relevance to the ML outcomes being predicted. For instance, some studies have successfully leveraged scores from multiple questionnaires, such as the Diagnostic and Statistical Manual of Mental Disorders (DSM), Depression Anxiety and Stress Scale (DASS), Edinburgh Perinatal/Postnatal Depression Scale (EPDS) Center for Epidemiological Studies-Depression (CES-D) survey, Mean Opinion Score (MOS), Hamilton Depression Rating Scale (HAM-D), State-Trait Anxiety Inventory (STAI), Posttraumatic Stress Disorder Checklist for DSM (PCL), Beck Depression Inventory (BDI, Beck Anxiety Inventory (BAI), Hospital Anxiety and Depression Scale (HADS), Goldberg's Depression Scale (GDS), self-reports and clinician reports, [94], [95], [96], [97], [98], [99], [100], [101], [102], [103], [104], [105], [106], [107], [108], [109], [110], [111].

## Preprocessing Techniques

In this section, important preprocessing techniques that have yielded significant findings and how they are used to help the detection of stress and its related MDs have been reviewed.

*Synthetic Minority Oversampling Technique (SMOTE) (n=3)*: In detection of stress and its related MDs, usually the number of samples for the stress or MD class is significantly lower than the non-stress or non-MD class. This imbalance in the number of samples for each class leads to a bias in prediction (towards the majority class). To correct for data bias, it is possible to oversample the underrepresented group. In stress detection studies using ML models, SMOTE is one of the most common approaches to boost the minority class using, which creates new samples by synthesizing those already available in the data (by combining their features) [77], [95], [112].

*Early Modality Fusion (n=1)*: In ML models used for prediction of stress with a multimodal approach, it has been shown that early fusion of multimodal data before feature extraction is more effective and archives a better performance. This is due to the fact that early modality fusion catches better the important characteristics that are in coherence with each other. For example a study showed that combining different measures including skin response, skin temperature and body acceleration before feature extraction outperforms the approach that extracts the features for each measure separately and combines them afterwards (Figure 4) [113].

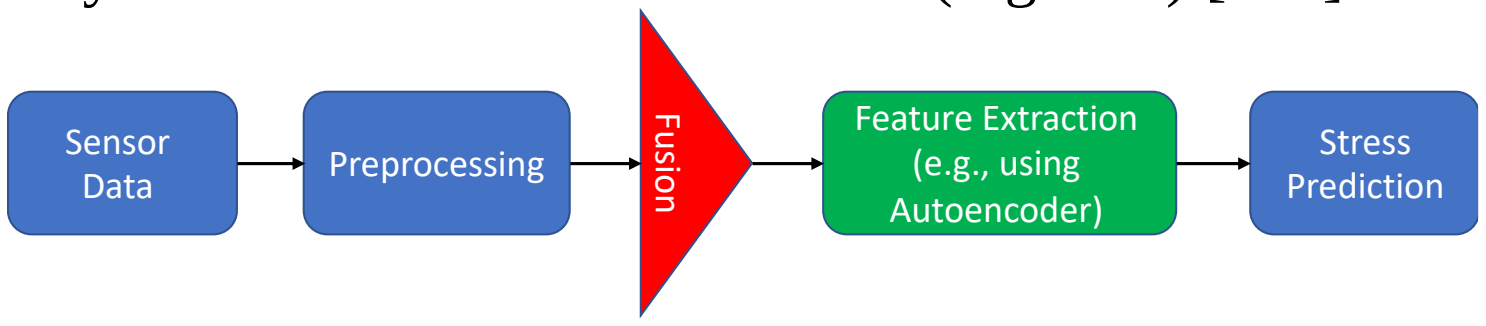

(a)

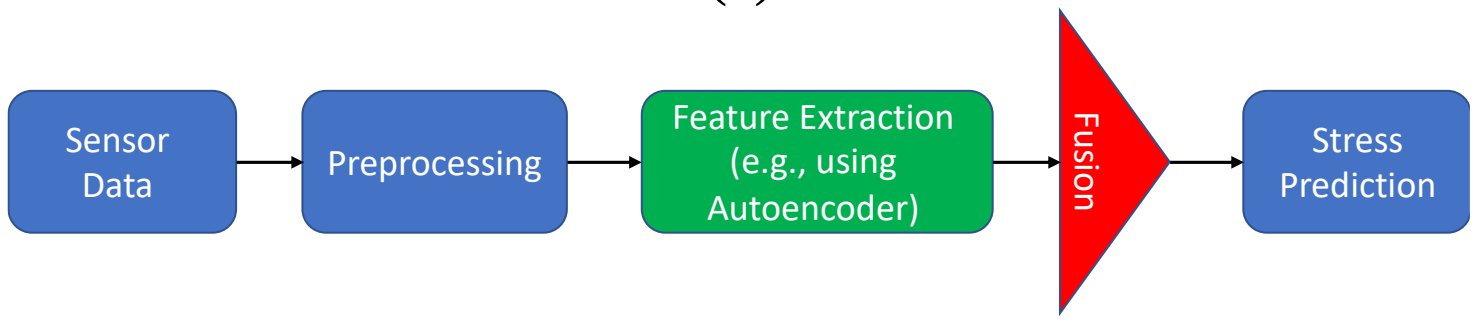

(b)

**Figure 4.** (a) Early Modality Fusion (b) Late Modality Fusion

*Power Spectral Density (PSD) (n=13)*: In physiological signals for stress detection, usually power of the signal changes during the moments of stress. PSD explains the frequency-based power distribution of a time series and reveals the locations of strong and weak frequency variation. Welch's method is one of the most common approaches to calculate PSD [49]. PSD is often used in the studies that include frequency domain HRV features for stress detection such as total HF or LF power [66], [67], [68], [86], [114], [114], [115], [116], [117], [118], [119], [120], [121].

*ILIOU (n=1)*: In detection of MDs such as depression and anxiety using machine learning techniques, having the least error rate is significantly important so that the person can take further actions appropriately. In this matter, data preprocessing step has an important role to minimize the noise and bias towards the false prediction. Iliou et al. proposed ILIOU, a data mapping and transformation method, that identifies useful information for detection of MDs, especially for depression. This method outperforms common data preprocessing techniques such as Principal component analysis (PCA), Evolutionary Search Algorithm (ESA) and Isomap for detection of depression [99].

*Principal Component Analysis (PCA) (n=3)*: Principal component analysis (PCA) is a method for lowering the dimensionality of such datasets while maximizing interpretability and minimizing loss of information. It does this by generating new variables that are uncorrelated and progressively optimize variance [42], [108], [122].

*Independent component analysis (ICA) (n=4)*: Independent component analysis (ICA) is a computational and statistical method for uncovering hidden elements underlying random variables, observations, or signals. This method is mostly used for removing artifacts from stationary signal noises of the multi-channel data. ICA optimizes higher-order statistics such as kurtosis, while PCA optimizes the covariance matrix of the data, which reflects second-order statistics. In stress detection using physiological signals that contain stationary noises (e.g. eyeblink noise in EEG) it is recommended to remove noises using ICA [47], [48], [49], [51].

*Artifact subspace reconstruction (ASR) (n=1)*: ASR is an adaptive approach for removing artifacts from signal recordings online or offline, mostly non-stationary signal noises. To identify artifacts based on their statistical qualities in the component subspace, it repeatedly computes a PCA on covariance matrices [123]. Since there are usually lots of non-stationary noises in the EEG data, in order to classify stress in multiple levels using EEG data, using ASR before classification is highly recommended [49].

*Latent Growth Mixture Modeling (LGMM) (n=1)*: Growth mixture modeling (GMM) is to discover numerous hidden subpopulations, describe longitudinal development within each hidden subpopulation, and investigate variation in hidden subpopulations' rates of change. Latent growth mixture models are gaining popularity as a statistical tool for estimating individual development over time and for probing the presence of latent trajectories, in which people belong to trajectories that are not directly observable [46], [124], [125].

*Dynamic Time Warping (DTW) (n=1)*: It is common practice to transform data from two time series into vectors and then compute the Euclidean distance between the resulting points in vector space to determine the degree of similarity or dissimilarity between the series, regardless of if they vary in time or velocity. DTW method can be applied to find such similarities that may exist between people in terms of their mood series. As an example, one may compare time-series to find whether they match for stress, depression, or anxiety. Moreover, it can be utilized to forecast the mental condition of persons with substantially comparable series patterns [115], [126].The difference between DTW and Euclidian matching is that unlike Euclidean matching, DTW considers distance of each point in one sequence, to every point in the other sequence to determine the similarity between them (Figure 5).

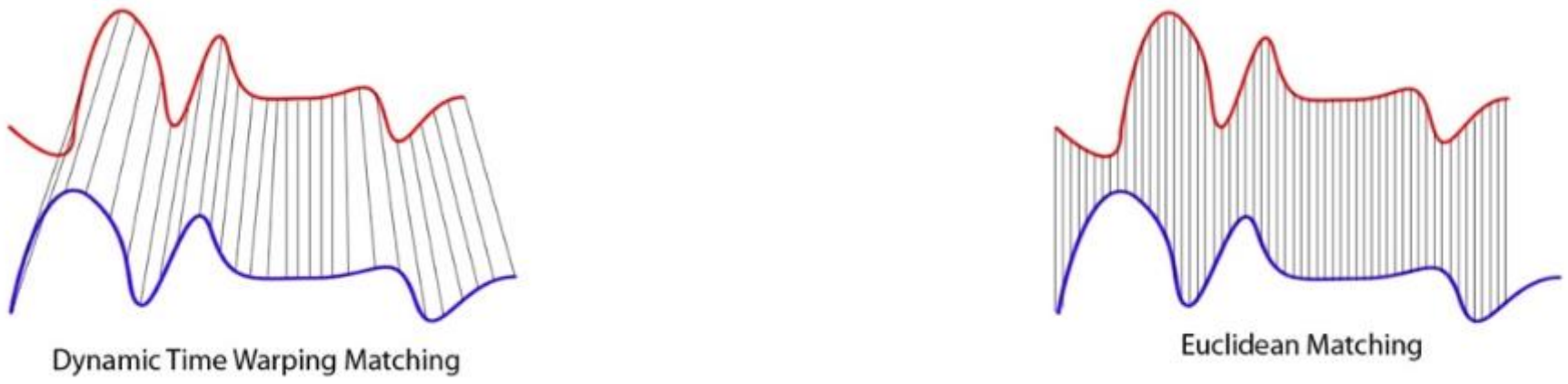

**Figure 5.** Dynamic Time Warping Vs. Euclidian Matching [127]

*Kalman Filter (n=2)*: The Kalman filter is a technique for making predictions about unknown variables (e.g., missing data) based on observable data. Kalman filters include two iterative steps—predict and update—that are used to estimate states using linear dynamical systems in state space format. Iterative cycles of predict and update are performed until convergence is achieved [128]. Kalman filter has been used to handle the missing data for stress detection in some studies [129], [130].

*Autoencoders (n=3)*: Autoencoders are a type of Neural Networks that learn a representation of the data in lower dimensions than the original data (encoding) by regenerating the input from the encodings (decoding). For data with very high dimensionality, usually clustering is not optimized because of the noise present in the original data. Hence, it is an appropriate practice to use the encoded representation of the data, obtained by autoencoders, to have lower and more optimized dimensions for clustering [49], [93], [131].

*Self-Organizing Map (SOM) (n=3)*: In ML, a self-organizing map (SOM) produces a low-dimensional – typically two-dimensional – representation of a high-dimensional dataset while preserving its topology by creating clusters. It is therefore possible to visualize and analyze high-dimensional data more easily (Figure 6) [92], [118], [132].

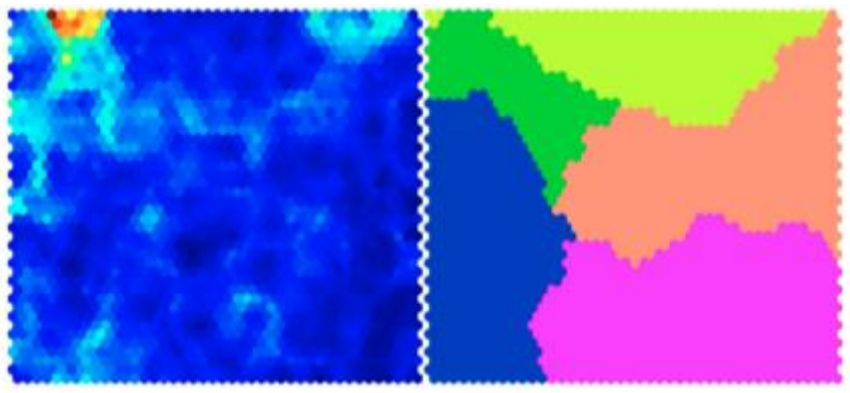

**Figure 6.** Representation of SOM before (left) and after mapping (right) [118].

*Wrapper Feature Selection Methods*: Wrapper methods try to use a subset of features while training a model. Changes will be made to the feature subset based on the performance about the prior model (Figure 7). Therefore, finding the best features using wrapper method is a search problem. These methods often have high computing costs [133]. Some most common wrapper methods are: Naïve search, Sequential Forward Feature Selection (SFFS), Sequential Backward Feature Selection (SBFS), and Generalized Sequential Search (GSS) [134]. Some studies used this approach as their feature selection technique [56], [59].

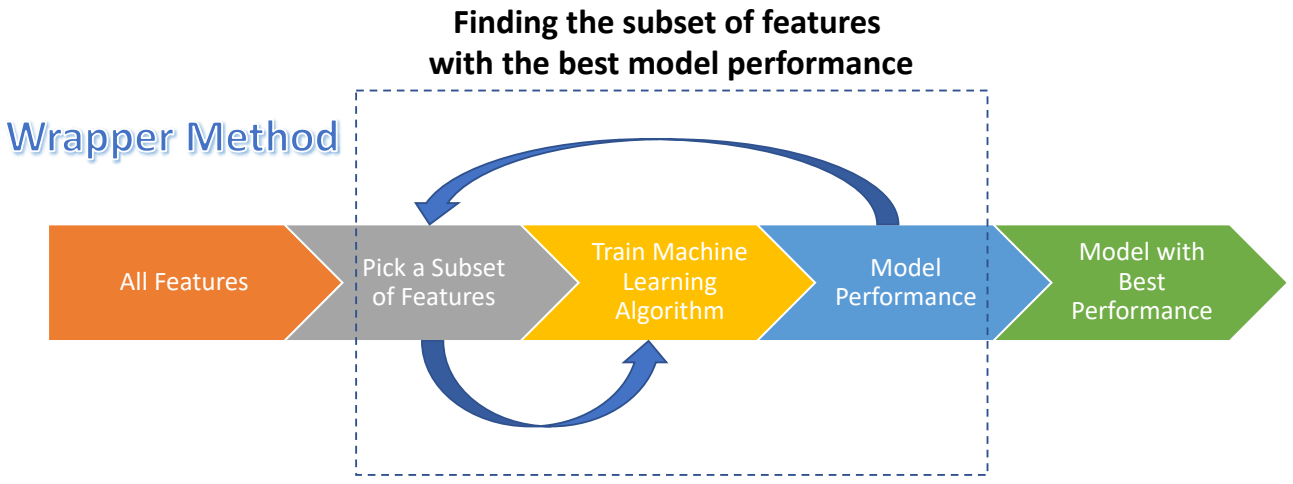

**Figure 7.** Steps of a wrapper feature selection method

*Filter Feature Selection Methods*: In general, filter methods are used as a preprocessing step without regard to any ML algorithms. Statistic tests are used instead to select features based on their correlation with dependent variables (Figure 8). The filter feature selection methods used in the literature are mentioned below.

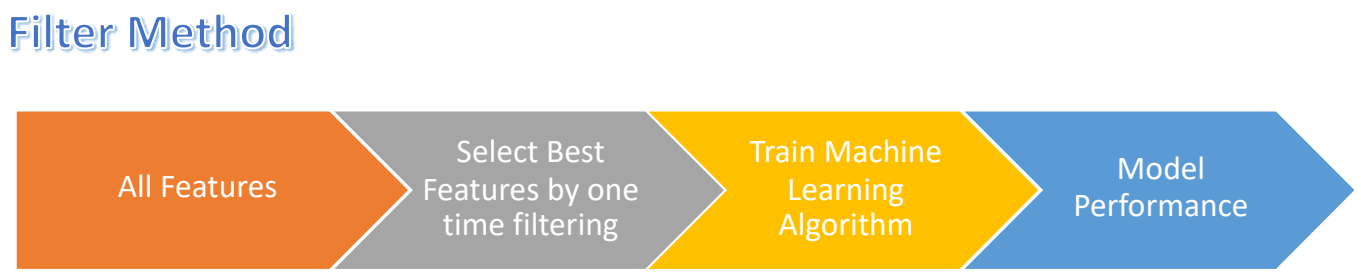

**Figure 8.** Steps of a filter feature selection method

- *Chi-square test (n=3)*: This test checks for independence between categorical features and the target variable. Features with high Chi-square scores are selected, implying a strong association with the target variable, which may be valuable for the model [40], [120], [135].

- *Pearson Correlation (n=2)*: Pearson linear correlation coefficient is a way to quantify how closely two sets of data are correlated linearly. It indicates how different measures are related to each other by a number between -1 to 1. Therefore, among highly correlated variables some them can be removed as they don't add useful information to ML models [98], [136].

- *Minimum Redundancy Maximum Relevance (mRMR) (n=2)*: mRMR technique chooses characteristics having a high correlation to output (relevance) and a low correlation to one another (redundancy). F-statistic is used to determine the correlation between features and the output, whilst Pearson correlation coefficient (for non-time series features) and Dynamic Time Warping (DTW for time series features) may be used to calculate the correlation between features (Figure 9). The objective function, which is a function of relevance and redundancy, is then maximized by selecting features one at a time using a greedy search. Mutual Information Difference (MID) and Mutual Information Quotient (MIQ) criteria are both frequently employed objective functions that depict the difference or quotient between relevance and redundancy [137], [138]. Using this feature selection method, Giannakakis et al. have ranked ECG measurements in the order of importance as mean HR , LF, NN50, standard deviation of HR,pNN50, LF/HF, RMSSD, HF, and total power [115].

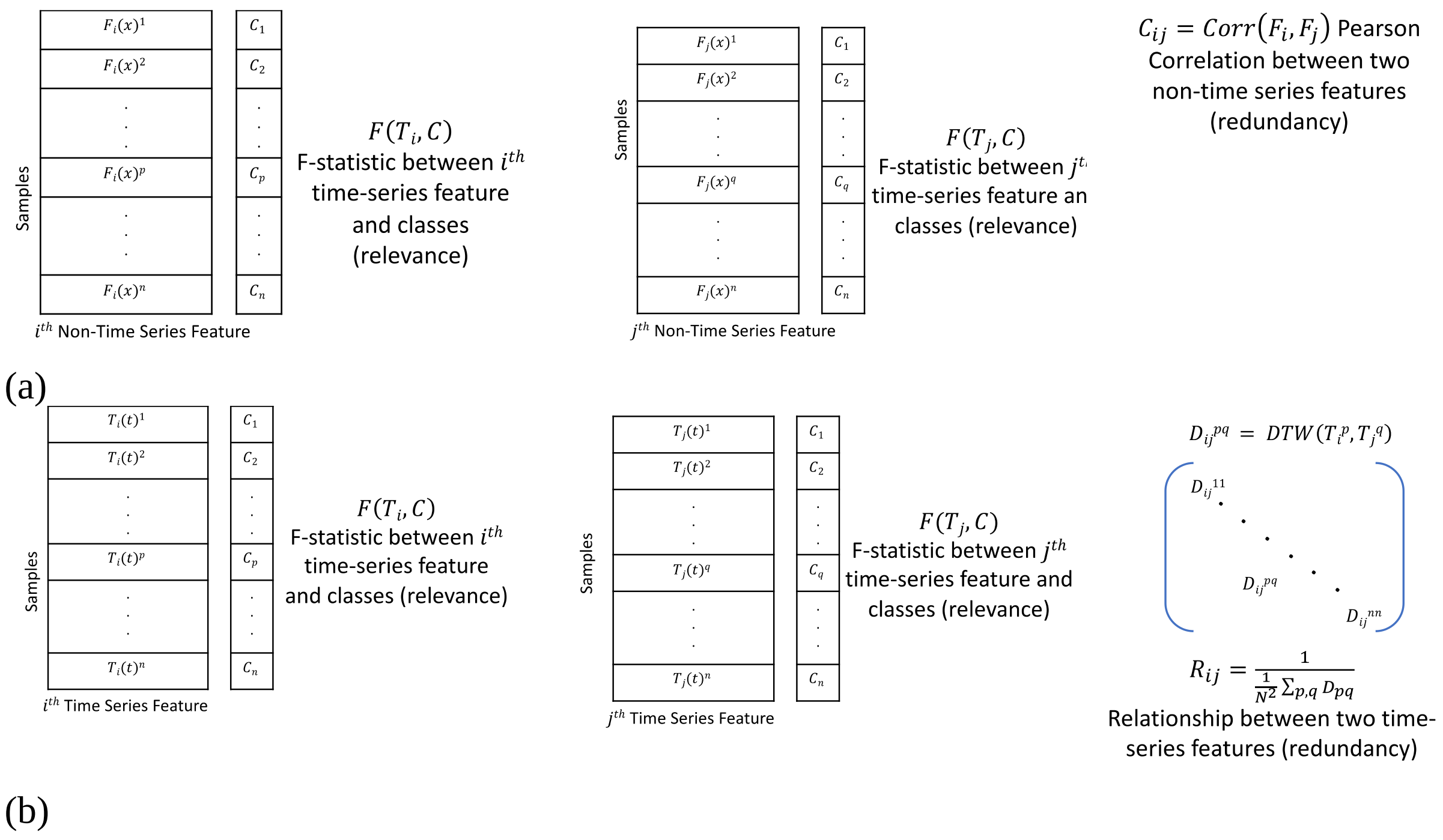

**Figure 9.** calculation of relevance and redundancy for (a) non-time series features (b) time-series features
(DTW: Dynamic Time Warping)

## Machine Learning (ML) Techniques

The ML algorithms used for stress and MD detection have been reviewed in this section. The papers used DL approach or Neural Network (NN, n=39) Logistic Regression (LR, n=26) Naive Bayes (NB, n=22), Decision Tree (DT, n=23), Boosting (e.g., Adaptive Boosting (AdaBoost), extreme Gradient Boosting (XGBoost), etc., n=22), Random Forest (RF, n=36), Discriminant Analysis (e.g., Linear Discriminant Analysis (LDA), Quadratic Discriminant Analysis (QDA), n=6), Fuzzy *C-means* (n=2), K-nearest neighbors (KNN, n=22) and Support Vector Machines (SVM, n=48). Figure 10 shows the distribution of articles by ML model. Refer to table A1 to find which papers have used each ML technique.

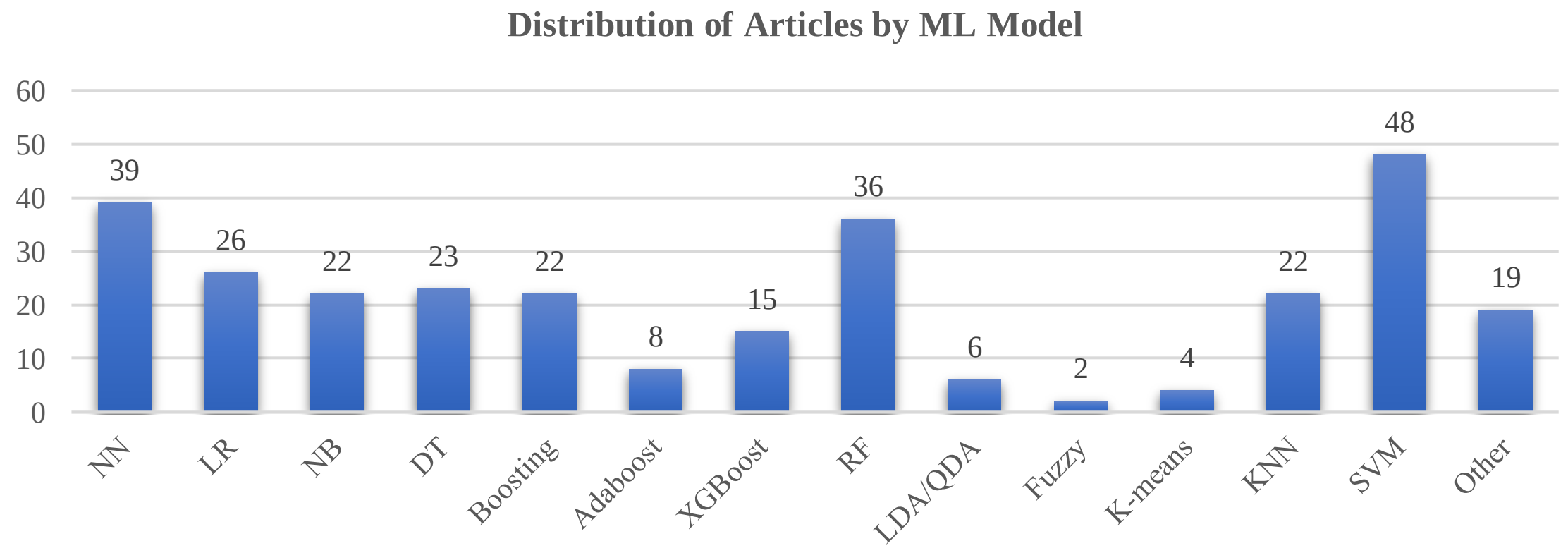

**Figure 10.** Number of articles for each ML model

*Logistic Regression (LR) (n=26):* LR is a supervised parametric ML technique in which multiple independent variables will be utilized to detect the occurrence of stress or normal condition [56], [102]. Some studies utilized the numerical independent variables (e.g., HRV time-domain features: RMSSD, HR, pNN50) [79], [139] or categorical data (e.g., answer to multiple choice questions) obtained from questionnaires [92], [99], [100].

*Naïve Bayes (NB) (n=22):* Naïve Bayes algorithm is a supervised, generally parametric, classification method that uses the Bayes Theorem as its foundation and has the naïve assumption of predictor independence. In other words, Naïve Bayes classifier assumes that the existence of a given independent variable to predict the dependent variable is independent of the presence of any other independent variable that predicts the dependent variable.

*Decision Tree (DT) (n=23):* Decision Tree is a supervised non-parametric ML algorithm used in classification and regression applications. It comprises a root node, branches, internal nodes, and leaf nodes in a hierarchical, tree-like structure (Figure 11).

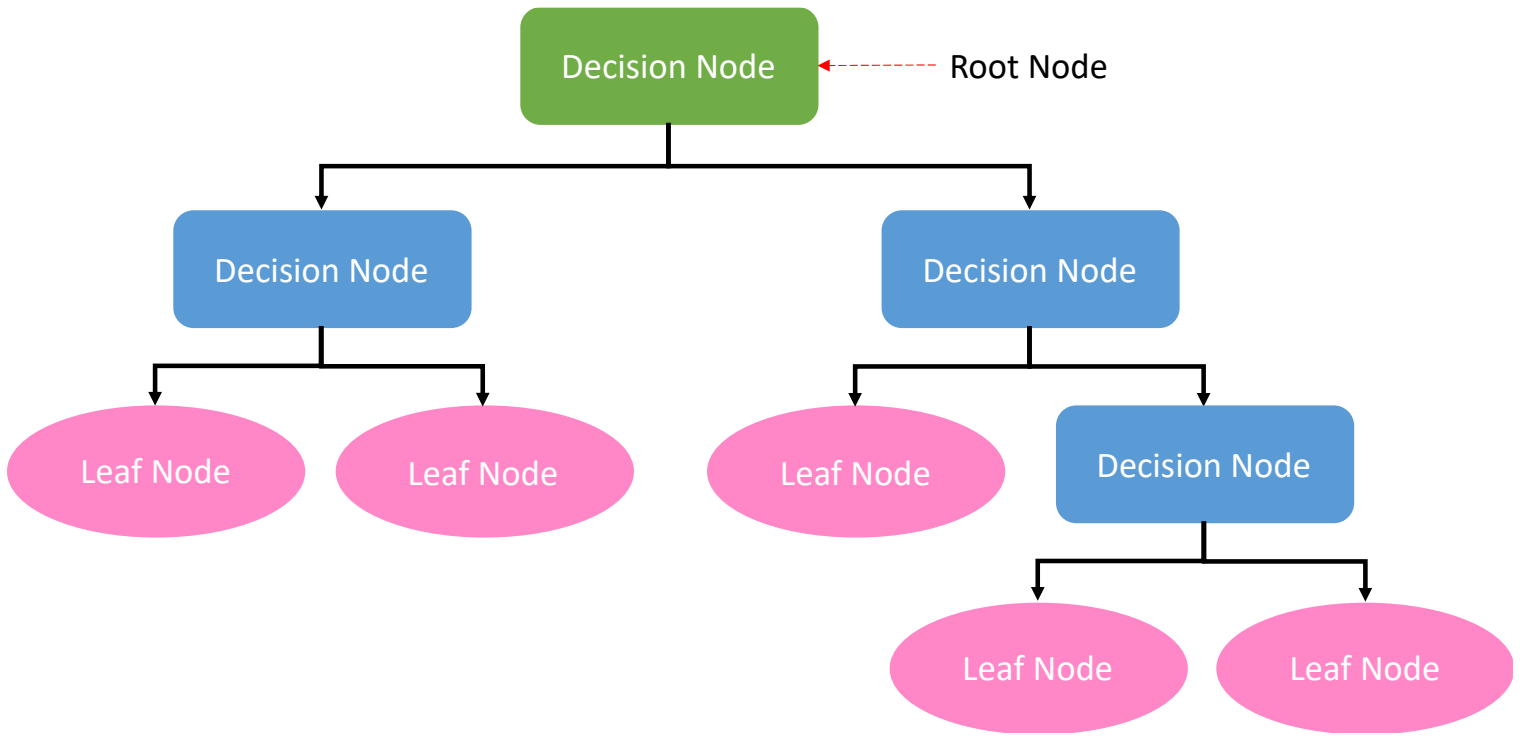

**Figure 11.** Structure of a Decision Tree

*Boosting (n=22):* Boosting is an ensemble learning for reducing training errors by combining a group of weak learners. When using Boosting algorithm, models are fitted on random samples of data, and then models are trained repeatedly in a sequence. When each model starts being trained in that sequence, it attempts to make up for the flaws of the one that came before it. The most commonly used Boosting algorithms are: Adaptive boosting (AdaBoost), Gradient boosting, and Extreme gradient boosting (XGBoost).

*Random Forest (RF) (n=36):* Random Forest is a supervised non-parametric ensemble learning algorithm that uses many Decision Trees built during the training process. Random Forest algorithm is used for both classification and regression problems. When it comes to classification, the Random Forest's output is the class that the majority of the Decision Trees choose. For regression purposes, an individual tree's predicted mean or average is returned as the output. Using Random Forests, we can overcome the tendency of decision trees to overfit to their training data.
*Discriminant Analysis (n=6):* Discriminant Analysis is a supervised parametric classification algorithm that works with data including a dependent variable and independent variables and mostly used to classify the observation into a certain group based on the independent variables in the data. Linear Discriminant Analysis (LDA) and Quadratic Discriminant Analysis (QDA) are the two forms of Discriminant Analysis.

*K-nearest neighbors (K-NN) (n=22):* K-nearest neighbors (K-NN) is a non-parametric supervised ML algorithm that is used for both classification and regression purposes. In classification, the algorithm determines the label of a new sample not available in the training data by assigning the label of the majority of k-nearest training data points to that new sample (Figure 12). In regression, the output for each sample, is the average of the values of k-nearest neighbors to that sample (not including the sample itself). In this literature K-NN has been only used for classification.

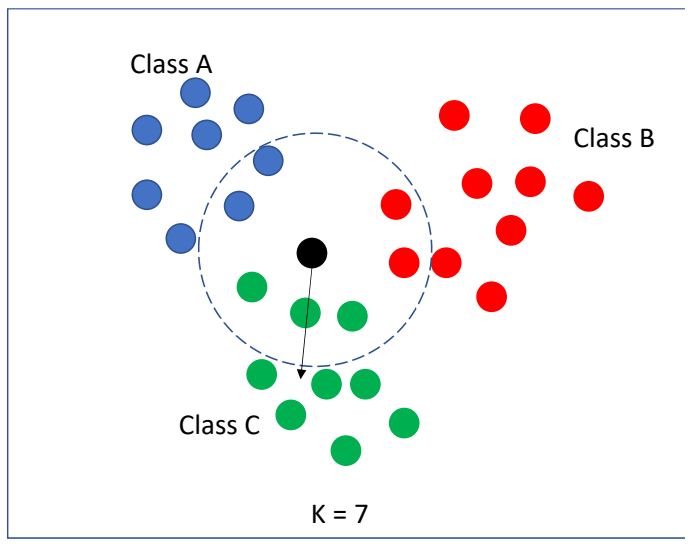

**Figure 12.** Example of K-NN classification with K = 7. In this example, the label of “Class C” is assigned to the new (black) datapoint since the majority of the 7-nearest datapoints to the new datapoint are from “Class C”.

*Support Vector Machines (SVM) (n=48):* Support Vector Machine is a parametric supervised ML algorithm used for both classification and regression problems. It can solve both linear and non-linear problems using non-linear kernels. For classification, the SVM algorithm finds a line (or a hyperplane for non-linear kernels) between each pair of classes of the training data in a way that the margin distance of that line or hyperplane to the closest point of each of those two classes is maximized (Figure 13). This is repeated for all pairs of classes in the dataset. Then the obtained lines are used as boundaries for classes. In regression, the SVM tries to find the line/hyperplane that within a very small margin of $\varepsilon$ (epsilon) has maximum number of datapoints. That line/hyperplane used for regression.

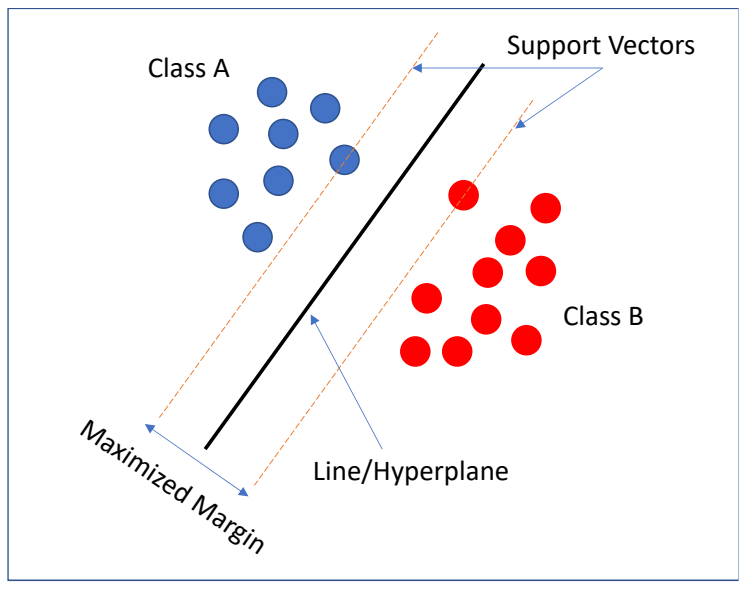

**Figure 13.** Visual representation of Support Vector Machine algorithm

*K-means clustering (n=4):* K-means clustering is an unsupervised ML algorithm that aims to arrange objects into groups based on their similarity. To find those similarities, it calculates the distance of data points into K random cluster centroids and assigns each data point to its closest centroid. Then location of each centroid is then updated by average value of all datapoints associated with that centroid. This process is repeated until there is no change in the location of the centroids. In ML models for stress detection, K-means clustering has been used in the literature for personalization of the ML models [42], [51], and for labeling the dataset [131], [140].

*Neural Network (NN) (n=39):* DL methods are a subset of ML methods that. NNs are the heart of the DL algorithms. The neural network is a method for implementing ML that utilizes interconnected nodes or neurons arranged in a layered structure resembling the human brain. There are different types of NNs have been explained below:

- *Artificial Neural Network (ANN):* It is possible to think of a single perceptron (or neuron) as an abstract Logistic Regression. In each layer of ANNs, a group of multiple perceptron or artificial neurons is used. Figure 14 shows an ANN with one layer and its working mechanism.

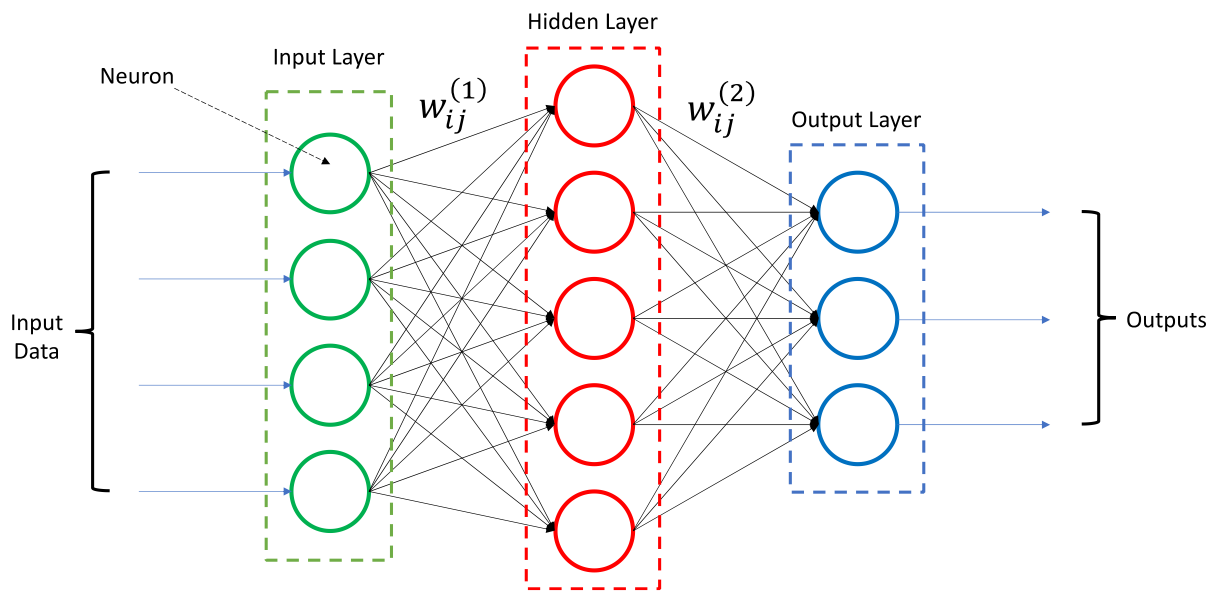

(a)

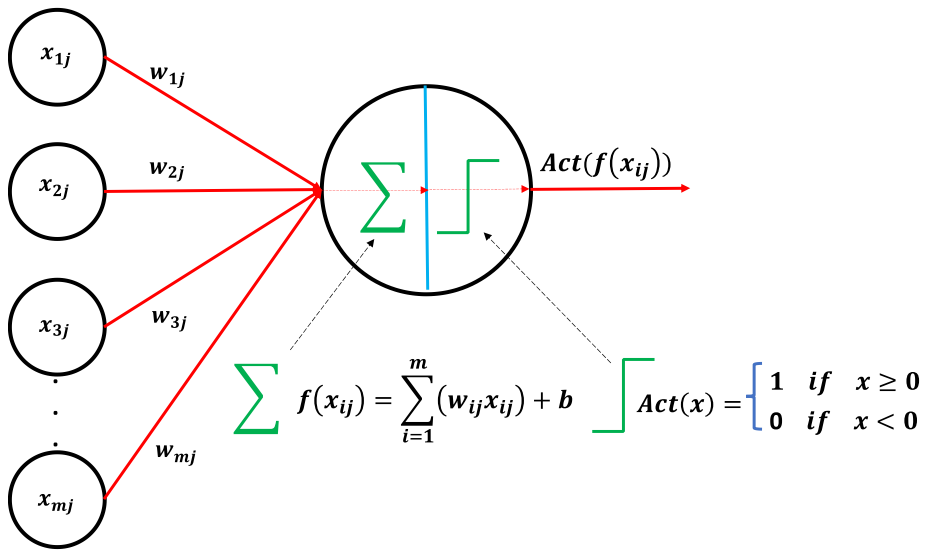

(b)

**Figure 14.** (a) Representation of an ANN with one hidden layer. $W_{ij}^{(1)}$ and $W_{ij}^{(2)}$denote the weights of the links connecting the first layer (input layer) to the hidden layer and weights of the links connecting the second layer to the next layer (output layer), respectively. (b) Representation of how a single neuron works. First, all the outputs of the previous layer are multiplied by the weights associated with the links connecting them to the $j^{th}$ neuron of the next layer and summed by a bias (summation and bias step). The result is then passed through an activation function (activation step).

- *Convolutional Neural Network (CNN):* CNNs are a form of neural network that are especially adept at handling data structures with a grid-like layout, such as images/objects. Classification and computer vision applications are common uses for convolutional neural networks (ConvNets or CNNs) (Figure 15).

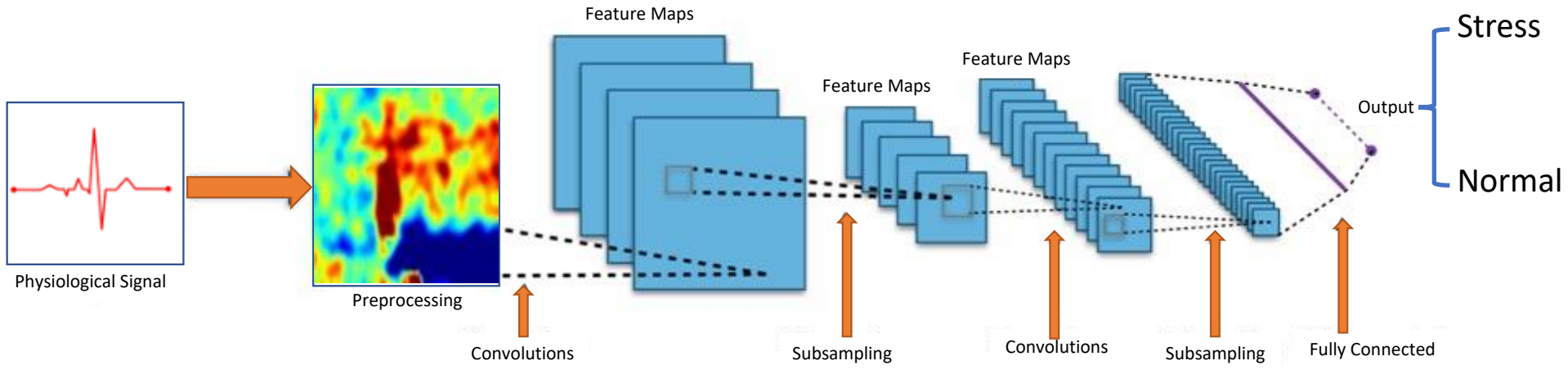

**Figure 15.** Representation of CNN for physiological signal

- *Recurrent Neural Network (RNN):* An RNN is a subset of artificial neural networks designed specifically for use with time series data and other sequence-based data. Long Short-Term Memory (LSTM) networks are the most common type of RNNs. In RNNs, Attention mechanism is a method that simulates cognitive attention in neural networks. The purpose of the impact is to encourage the network to give greater attention to the small but significant portions of the input data by enhancing some and reducing others. Since stress may alter a small portion of physiological data (e.g. ECG), attention mechanism can be used to detect stress using RNNs when large datasets are available [141].

Cong et al. introduced X-A-BiLSTM, which is a DL model that includes XGBoost (to filter data and handle imbalanced data) and Attention Bi-LSTM (LSTM with forward and backward memory and Attention mechanism) Neural Network used for stress classification using text data [87].

*Other ML techniques (n=19):*

- *Voting ensemble classifier:* The classification is decided based on weighted voting, which is determined by using a voting ensemble approach. The voting classifier allows for voting in which the final class labels are determined either by the class chosen most frequently by the classification models, or by the average of the output probabilities from each classification model. In the literature, this method has been utilized for PTSD detection [112], stress and stress related MDs [68], [80], [103], [140], [142].

- *Fuzzy C-means (FCM) clustering:* Fuzzy C-means clustering (FCM) is a clustering approach that assigns every data point to all the clusters with a certain probability instead of assigning each point to only one cluster. A data point that is near to the cluster's center, for instance, will have a high degree of membership there, while a data point that is distant from the cluster's center would have a low degree of membership [143]. Since depression, anxiety are not discrete measures, some studies have used FCM as an alternative to other clustering techniques for detection of these MDs [99], [101].

In this article, the recent ML algorithms, preprocessing techniques, and data (e.g., physiological data, questionnaire data, etc.) used in detection, prediction and monitoring of stress and the most common MDs (i.e., depression, anxiety, other stress-related MDs) have been reviewed.
Based on this review, it is concluded that among classic ML algorithms (excluding DL approaches), supervised models of Support Vector Machines (SVMs) and Random Forest (RF), have been used more often and achieved better performance in terms of model accuracy and robustness (measured by parameters like Area Under the Receiver Operating Characteristic curve (AUROC)). The accuracy of ML models is a critical indicator of their utility in real-world applications. The review demonstrates that SVM consistently achieves high accuracy across various data types, including HR, HRV, and skin response. For instance, SVM achieved 93% accuracy with HR, PPG, and skin response data in study [34], and 96% with skin response data in study [140]. These results underscore SVM's robustness in handling complex, non-linear data. Random Forest also shows commendable performance, with an accuracy of 99.88% in study [144], reflecting its strength in ensemble learning to mitigate overfitting and noise.
Moreover, among the predicting measures for stress and stress-related MDs, HR, HRV and skin response have been used the most often (Figure 16). These measures were the major explaining factors in the ML algorithms to predict stress and stress-related MDs. It is noticeable that DL approaches are becoming more popular as these techniques provide unique specifications that classic ML algorithms cannot provide.

Since stress is a time dependent event, the relationship between different lags of time can be important for detection of stress. Recurrent Neural Networks (RNNs) and Convolutional Neural Networks (CNNs) will take into account the relationship between datapoints in different time-series for their decision making and they have the potential to enhance the detections. Deep learning models, specifically CNNs and LSTMs, show promising results, with CNNs achieving 92.8% accuracy in HRV and ECG data in study [139], indicating their potential in feature-rich physiological data. However, it is worth noting that deep learning models require substantial data for training, which may limit their applicability in studies with smaller datasets. Attention mechanism in RNNs is a new technique that is becoming popular for finding animalities in physiological signals. However, based on the review of literature, this mechanism has only been used on text data (not on physiological signals) to detect stress. Therefore, Attention mechanism is technology that can be further utilized for physiological signals to detect stress.
Unsupervised ML (and DL algorithms) such as clustering techniques have been used mostly for the preprocessing step to label the data (if labels are not available) and also for finding a representation of the data that achieves the best performance in detection algorithms.

For data preprocessing, feature selection (i.e., filter and wrapper methods) and extraction techniques have been commonly used. In feature extraction approaches, latent representations of data by transformations such as output of encoder in autoencoders have been useful to remove data noises and to make the data more compact, making further computations more efficient. PCA and ICA are other most common feature extraction approaches used in the literature.

Among the selected features, statistical indicators of heart measurements such as mean, standard deviation of HR, along with time and frequency representations of HRV such as RMSSD and total LF and HF power were most widely used. Heart measurements also have been more often than other measurements as they are unobtrusive, non-invasive, affordable and easier to measure and also describing a big portion of stress events. After those, skin response measure has been found as one of the most important factors in detection of stress and its related disorders. The time-frequency approaches to analyze time series data are getting more popular in this area as they are proper representations of data for DL approaches which can be more accurate and robust. As an example, for DL algorithms, RNNs with attention mechanisms can help to find portions of data related to stress and its related disorders with higher confidence.

Most of the studies models do not interpret the ML models and look at them as black box. This limits the contribution to the body of science. SHapley Additive exPlanations (SHAP) is a technique used by some studies to interpret the models such as evaluation of features to find the most important ones and also how in what direction each feature affects the predictions. SHAP correlation plot provides insight into the distribution of the features themselves, as well as the relationship between their influence on the model. In other words, it provides the importance of each feature on prediction of the dependent variable by taking into account both the main effect as and the interaction effect of that feature with other features in the data [46], [77], [105], [144], [145], [146].

Despite progress in stress detection methodologies, the exploration of personalized models has been limited. Most studies have not gone beyond basic normalization techniques, overlooking the fact that physiological measures are as distinct to individuals as biometric identifiers. A notable exception can be found in a select few studies [51], [113], [147], which have employed more sophisticated personalization techniques, integrating complex data transformations to account for individual variability.

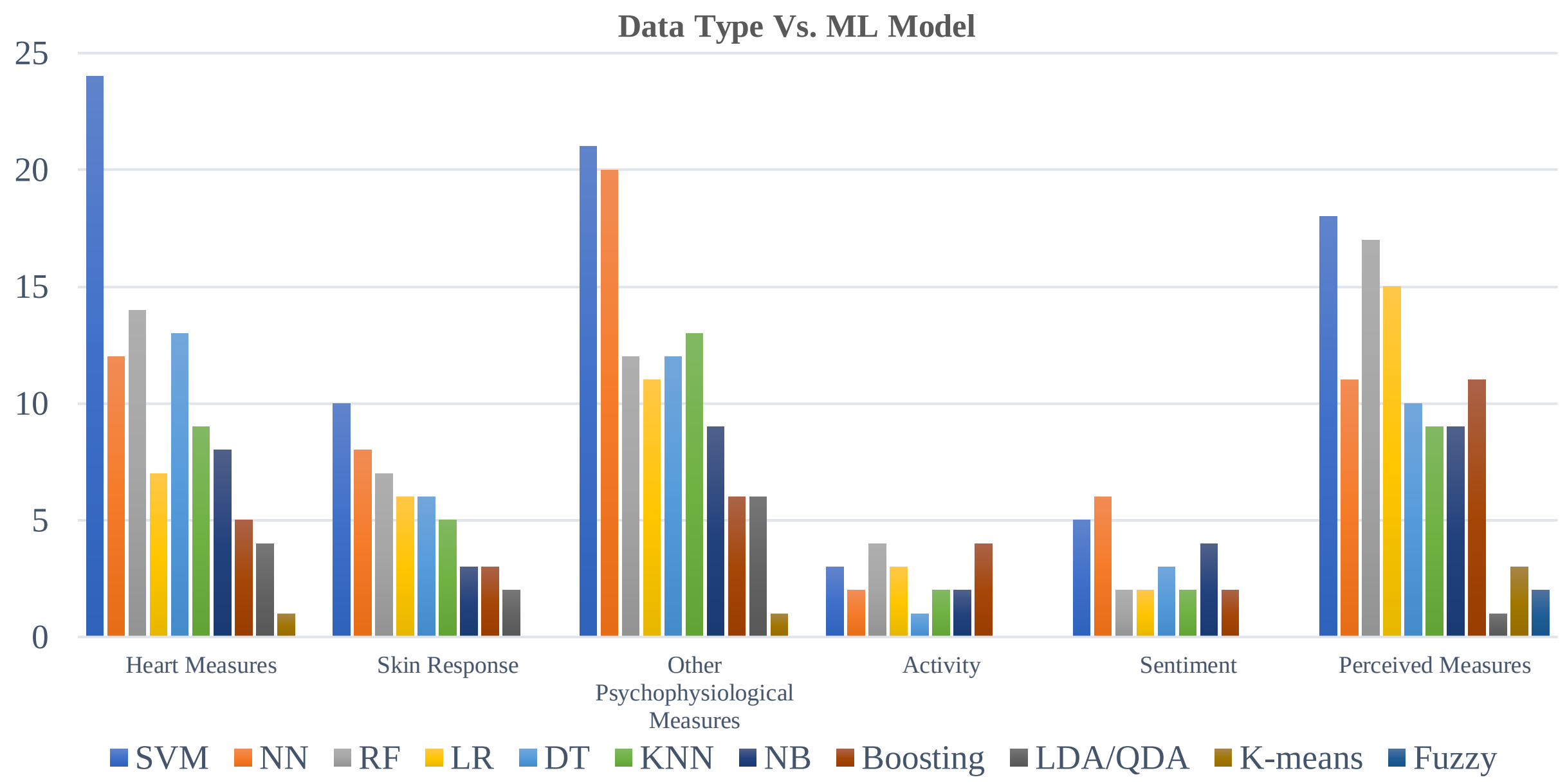

**Figure 16.** Distribution of ML models used for each type of data. In this figure, **skin response** and **heart measures** (including HR, HRV and blood pressure) have been shown separately due to their high usage and importance in the literature. **Other psychophysiological measures** include EEG, EMG, Eye-tracking and respiratory signals. **Activity** includes body movement. **Sentiment** data includes speech and text data. Finally, perceived measures include questionnaire and self-report data.

## Strengths of the review

In undertaking this scoping review, we have embarked on a rich exploration of the applications of machine learning (ML) in the field of stress detection, articulating a narrative that is both comprehensive and detailed. The review lays out a landscape where diverse data types are not merely cataloged but deeply analyzed for their roles and interconnections within the broader context of methodological approaches. This provides a robust understanding of the field's current state and its complexities.

This review has documented a comprehensive assessment on various physiological measurement techniques, including heart rate variability (HRV), electroencephalograms (EEG), and electrocardiograms (ECG), etc. This assessment is not just a recounting of the types of data employed in the literature but a thoughtful consideration of how each contributes to a multifaceted understanding of stress indicators. It is an acknowledgment that the signals of stress are as complex as the condition itself, necessitating a rich palette of investigative tools.

The review also examines a range of advanced preprocessing techniques such as mRMR, SOM, SMOTE and PCA. This examination sheds light on how different studies leverage these methods to refine the quality of data fed into ML models, thereby potentially enhancing the models' accuracy and reliability in detecting stress. It is an illustration of how sophisticated data treatment can lead to more nuanced insights, even if our own methodology did not directly employ these techniques.

### Limitations

Our scoping review acknowledges its inherent constraints, including a possible selection bias due to potential omissions of pertinent studies. It serves as a contemporary cross-section of the rapidly evolving domains of machine learning and mental health, underscoring the imperative for periodic scholarly review to sustain its relevance and precision. While we survey a broad spectrum of machine learning techniques applied to stress detection, we do not extensively assess their efficacy, suggesting a fertile ground for future empirical investigations to assess these methods across diverse data cohorts and settings. Additionally, while we address the preprocessing techniques and their impact on model performance, our discussion does not delve into detailed technical analysis. Finally, the crucial issue of model interpretability is touched upon but not explored in depth, presenting an opportunity for further scholarly explorations.

## Conclusions and Future Directions

The pivotal insights from this review underscore the potential of ML to redefine the approach to mental health care, particularly in the diagnosis and management of stress-related conditions and MDs. As we have discerned, there is an expansive field ripe for further exploration, with research gaps suggesting a number of promising directions. Guided by these insights, we can now chart a course for future research that not only expands the boundaries of our scientific understanding but also translates into tangible improvements in clinical practice.

*Real-time and Naturalistic ML Applications*

The scarcity of real-time studies in naturalistic settings has highlighted the importance of developing ML models that accurately reflect and respond to the complexities of real life. Future research must prioritize the creation of algorithms capable of operating amidst the unpredictability of daily life, providing immediate insights and adaptable interventions. These models hold the potential to transform practice by offering tools that can preemptively identify stress and MD symptoms, enabling clinicians to intervene before conditions worsen.

*Temporal Data and Deep Learning*

Our review illuminates the untapped potential of time series data in capturing the evolution of stress and MDs. Deep learning techniques, specifically designed to interpret complex, sequential data, could lead to breakthroughs in how we understand and predict mental health trajectories. For practice, this means more sophisticated diagnostic tools that can provide a nuanced picture of a patient's mental health over time, enabling personalized treatment plans that are responsive to the patient’s changing condition.

*Personalization in ML Models*

The need for individualized care in mental health cannot be overstated. The heterogeneity of stress responses and MD symptoms calls for personalized ML models tailored to individual physiological and behavioral patterns. Future research should focus on leveraging multi-task learning to refine algorithms that adapt to individual baselines, enhancing the personalization of care. For clinicians, this means access to tools that can more accurately reflect and respond to the unique needs of each patient, reducing the risk of misdiagnosis and improving treatment efficacy.

Predictive analytics can be instrumental in identifying key factors that contribute to misdiagnosis and delayed help-seeking. Future studies should look to build on this knowledge to inform the creation of interventions that encourage timely and accurate diagnosis. In practice, this could lead to the development of targeted screening tools that assist clinicians in recognizing at-risk individuals more effectively. The integration of clinical expertise with ML innovation is crucial for the development of tools that are both advanced and clinically relevant. Collaboration between healthcare professionals, patients, and AI developers will be essential in creating user-centered tools that address real-world needs. This collaborative approach will likely result in the development of AI applications that are more intuitive and effective in clinical settings.

***Conflict of Interest***

There is no conflict of interest.

## *Appendix*

**Table A1.** Summary of all articles included in the synthesis

| Article | Year | Data Type | Population Type | Population Demographic Data | Personalization Method | Real-Time? | ML Model | Data Collection Tool | Study Type | ML Type | Ground-Truth | Performance (Best Model) |
|---|---|---|---|---|---|---|---|---|---|---|---|---|
| **[40]** | 2017 | HR, PPG, Audio Signals | Children | | | Yes | NB, DT, SVM | Wearable Sensor, Microphone | Naturalistic | Classification | | Accuracy: 83% (DT) |
| **[97]** | 2018 | Questionnaire | Patients | N=270 | | No | Boosting, XGBoost | Questionnaire | Pre-collected data without experiment | Classification | DSM scores for depression | Accuracy: 97.8% (XGBoost) |
| **[88]** | 2020 | Text | College Students | N=693 (108 Depressed, 585 Normal) | | No | NN, Boosting, Adaboost, SVM | Weibo Sina Application | Naturalistic | Regression | frequency of positive and negative words | Precision: 0.88 (Adaboost) |
| **[104]** | 2018 | Questionnaire | Indian College Students | N=938 | | No | Ant Colony Optimization | DASS-21 Questionnaire | Lab | Classification | DASS-21 scores | Accuracy: 93.57% (ACO) |
| **[84]** | 2021 | Audio Signals | Depressed Patients and Healthy People | N=142 (42 Patient, 100 Healthy) | | No | NN | | Naturalistic | Classification | Depression Diagnosis (RSDD) dataset | Accuracy: 84% (CNN) |
| **[103]** | 2021 | Questionnaire | Unemployed people | N=334 (86 Depressed, 248 Normal) | | No | LR, NB, DT, KNN, SVM | | Naturalistic | Classification | Questionnaire scores for depression | Accuracy: 89.6% (Ensemble) |
| **[41]** | 2018 | HR, PPG, Acceleration/Body Movement, Skin Response | | N=6 | | No | NB, DT, SVM | Empatica E4 | Lab | Classification | Stress-inducing task | Accuracy: 93% (SVM) |
| **[148]** | 2018 | Skin Response, Questionnaire | | N=110 | | No | SVM | PCL-5 Questionnaire | Lab | Regression | PCL-5 questionnaire for PTSD | MSE: 138 (SVR) |
| **[71]** | 2020 | EMG, Skin Response | Young Gamers | N=15 | | Yes | NN | Wearable Sensor | Lab | Classification | Questionnaire scores | Accuracy: 95% (LSTM) |
| **[149]** | 2020 | HRV, ECG | | | | No | NN | BIOPAC MP150-BioNomadix | Lab | Classification | Three phases of driving task in driving simulator | Accuracy: 92.8% (CNN) |
| **[115]** | 2019 | HRV, ECG | | N=24 | Baseline Removal, Pairwise Transformation | No | NB, RF, LDA/QDA, KNN, SVM, Other | Ag/AgCl Electrodes | Lab | Classification | Stress-inducing task | Accuracy: 84.4% (SVM) |
| **[150]** | 2021 | Skin Response | Students | N=25 | Standard Scale on E-nose data. | No | KNN, SVM, LDA | Gas sensor | Lab | Classification | Stress-inducing task | Accuracy: 96% (SVM) |
| **[151]** | 2021 | HR, PPG, Skin Response | College Students | N=1 (22 years old) | | No | SVM | Empatica E4 | Lab | Classification | Stress-inducing task | Accuracy: 80% (SVM) |
| **[91]** | 2020 | Text | | ~11k Tweets | | No | LR, NB, RF, SVM | Twitter | Naturalistic | Classification | Labeling by sentiment analysis | Accuracy: 90% (SVM) |
| **[57]** | 2019 | Eye Tracking | | N=100 (50 Healthy, 50 Non-healthy) | | No | NN | Nvidia GPU | Lab | Classification | Number of rings in eye Iris | Accuracy: 98% (NN) |
| **[152]** | 2019 | Skin Response, PPG, HR, HRV | Patients during virtual speaking | N=30 | Min-max normalization, Zero means normalization | No | SVM | Empatica E4 | Lab | Classification | Stress-inducing task | Accuracy: 86.3% (SVM) |
| **[93]** | 2019 | Text | | ~14k posts from Weibo Sina | | No | NN, SVM, BERT | Weibo Sina Application | Naturalistic | Classification | Hamilton Depression Rating Scale for depression | F1 score: 0.538 (BERT) |
| **[106]** | 2019 | Questionnaire | Adult patients | | Normalization by Term Frequency–Inverse Document Frequency (TF-IDF) | No | NN, Other | | Naturalistic | Classification | Labels by clinician | Accuracy: 94.5%, AUROC: 98.5%(NN) |
| **[110]** | 2019 | Questionnaire | Adults above 50 | | | No | NN, LR, Boosting, XGBoost, RF | Questionnaire | Pre-collected data without experiment | Classification | CES-D scores | Log-loss: 0.241, AUROC: 0.886 (XGBoost) |
| **[67]** | 2022 | Respiratory, ECG, Questionnaire, HRV | Students | N=99 | Resting state ECG removed from data as baseline | No | NB, DT, LDA/QDA, KNN, SVM | E-prime software | Lab | Classification | STAI scores | Accuracy: 73.68% (DT) |
| **[140]** | 2018 | Questionnaire | Students and Working professionals | N=656 | | No | LR, NB, DT, K-means, KNN, SVM | Questionarate | Naturalistic | Classification | Clustering labels | Accuracy: 90% (RF) |
| **[53]** | 2019 | EEG | Students | N=63 | | No | KNN, SVM | PSS-14 Questionnaire, Neurosky Mindwave EEG | Lab | Classification | PSS-14 scores | Accuracy: 74.43% (KNN) |
| **[136]** | 2021 | HRV, ECG | Yoga and Chi practitioners | N=12 | | No | KNN | Precollected dataset (from PhysioNet) | Pre-collected data without experiment | Classification | Stress-inducing task | Accuracy: 95.31% (SVM) |
| **[50]** | 2019 | EEG, Skin Response, PPG, HRV | College Students and Instructors | N=28 | | No | NN, NB, SVM | MUSE EEG, Shimmer GSR+ | Lab | Classification | PSS scores | Accuracy: 75% (NN) |
| **[54]** | 2019 | EEG | College Students and Instructors | N=28 | | No | NN, NB, SVM | MUSE EEG | Lab | Classification | PSS scores | Accuracy: 92.85% (NN) |
| **[153]** | 2018 | Questionnaire, Text | Facebook users | N=431 (319 Depressed, 162 Normal) | | No | NN, Other | Facebook, CES-D Questionnaire | Naturalistic | Classification | CES-D scores | Accuracy: 72% (CNN) |
| **[131]** | 2019 | Questionnaire | College Students | N=48 | | No | NN, Boosting, XGBoost, RF, K-means, SVM | | Naturalistic | Classification | PHQ-9 scores | AUROC: 0.92 (SVM) |
| **[49]** | 2021 | EEG | Nurses and non-health professionals | N=80 (30 Nurses, 50 non-health professionals) | | No | Adaboost, RF, KNN, SVM, LDA, Ridge, Deep Bilief Network | NeuroScan EEG, SynAmps 2 amplifier | Lab | Classification | Self-report | Accuracy: 91% (SVM) |
| **[154]** | 2021 | HR, Skin Response, PPG | Engineering College Students | N=21 | | No | Boosting, Adaboost, XGBoost, RF, KNN | Empatica E4 | Lab | Classification | Stress-inducing task | Accuracy: 99.88% (RF) |
| **[79]** | 2021 | Acceleration/Body Movement, PPG, Questionnaire | Students and Staff | N=32 | PSS scores were used to change the threshold for stress detection | Yes | LR, Boosting, Adaboost, XGBoost, RF, SVM | Fossil Gen4 Explorist | Both lab and naturalistic | Classification | Stress-inducing task, Self-report | Accuracy: 82.6%, AUROC: 0.790, F-1 score: 0.623 (SVM) |
| **[80]** | 2020 | Acceleration/Body Movement, Skin Response, PPG, HR, HRV | Drivers | N=9 | One model was trained for each subject (with different thersholds for each subject) | No | RF | Empatica E4 | Pre-collected data without experiment | Regression | Baseline and driving condition | RMSE: 0.03 (Regression) |
| **[68]** | 2019 | Skin Response, Respiratory, ECG, HRV | Students | N=60 | | No | LR, NB, DT, RF, SVM, Other | thoracic respiration belt, skin conductance adhesive patches, BVP sensor | Lab | | | Accuracy: 74% (LR) |
| **[90]** | 2021 | Text | | | | Yes | NB, RF, KNN, SVM, Other | web page crawler | Naturalistic | Classification | Self-report | |
| **[155]** | 2020 | Questionnaire | Korean people | N=39,225 | | Yes | Context-DNN | | Pre-collected data without experiment | Classification | Precollected Depression Scores | Accuracy: 94.57% (Context-DNN) |
| **[129]** | 2019 | HR, PPG | Veterans | N=100 | | Yes | NN, NB, RF, SVM, Other | iPhone | Naturalistic | Classification | Self-reported PTSD triggers | AUROC: 0.67 (SVM) |
| **[116]** | 2019 | HRV, PPG, Acceleration/Body Movement, Skin Response | Students | N=21 | One model was trained for each subject | Yes | RF, KNN, LR, NN | Samsung Gear S, Empatica E4 | Naturalistic | Classification | PSS, NASA-TLX, STAI and other questionnaire scores | Accuracy: 97.92% (RF) |
| **[66]** | 2019 | HRV, Respiratory, ECG | Students | N=18 | Resting state removed from ECG and RESP data as baseline | No | NN | Zephyr BioHarness 3.0 | Lab | Classification | Stress-inducing task | Accuracy: 83.9% (DeepER) |
| **[48]** | 2019 | EEG | Healthy and Mild-depressed | N=22 | | No | Spiking Neural Network (SNN) | SynAmps amplifier, 61-channel EEG | Lab | Classification | BDI scores for depression | Accuracy: 72.13% (SNN) |
| **[117]** | 2020 | HRV, PPG, Questionnaire | Participants taking exams | N=632 | | No | NN | BioBeats Biobeam band | Naturalistic | Classification | PSS, STAI and DASS scores | Accuracy: 83% (LSTM) |
| **[85]** | 2019 | Video | | N=82 | | No | NN, Other | AVEC2013 and AVEC2014 datasets | Pre-collected data without experiment | Regression | BDI scores for depression | RMSE: 8.5 (CNN) |
| **[92]** | 2017 | Text | | 459 posts (207 distorted, 252 undistorted) | | No | NN, LR, NB, DT, KNN, SOM | | Pre-collected data without experiment | Classification | Hand-labeled data | Accuracy: 73% (LR) |
| **[118]** | 2017 | Skin Response, PPG, HRV | Healthy subjects | N=12 | | No | CNN, SOM | Samsung Gear VR | Lab | Classification | Stress-inducing task | Accuracy: 95% (K-ELM) |
| **[135]** | 2020 | Other | Women | N=69,169 | | No | NN, LR, DT, XGBoost, RF | | Pre-collected data without experiment | Classification | Depression from electronic health records (EHRs) | AUROC: 0.937 (LR) |
| **[113]** | 2020 | Acceleration/Body Movement, Skin Response | College Students | N=239 | Multi-task learning with participants as tasks in NN | No | NN | SNAP-SHOT dataset | Pre-collected data without experiment | Classification | Self-report | |
| **[56]** | 2020 | EEG | Healthy subjects | N=28 | | No | LR, RF, NN | MUSE EEG | Lab | Classification | STAI Scores | Accuracy: 78.5% (RF) |
| **[100]** | 2021 | Questionnaire | | | | No | LR, NB, DT, RF, KNN, SVM | Goldberg's Depression Questionnaire | Pre-collected data without experiment | Classification | Goldberg questionnaire scores | Accuracy: 86% (RF) |
| **[75]** | 2019 | HR, PPG, Acceleration/Body Movement, Skin Response, Other | Depressed and Healthy Japanese people | N=86 (45 Depressed, 41 Healthy) | | No | XGBoost | Silmee W20 Wristband | Naturalistic | Classification | healthy/patient participants | Accuracy: 76%, Sensitivity: 73%, Specificity: 79% (XGBoost) |
| **[119]** | 2019 | HRV, ECG | Firefighters | N=26 | | Yes | DT, SVM | Polar H7 chest strap ECG | Lab | Classification | Stress-inducing task | Accuracy: 88% (DT) |
| **[126]** | 2016 | Other | | N=270 | Normalizing the variables | No | RF, SVM | eMate EMA application, iYouVU application | Naturalistic | Regression | Mood self-report | MSE: 0.410 (SVM) |
| **[156]** | 2018 | ECG, HRV | Drivers | N=17 | | No | NN, DT, RF | Wearable Sensors | Pre-collected data without experiment | Classification | Stress-inducing task | Accuracy: Up to 100%, AUC: 1 (RF) |
| **[46]** | 2021 | HR, Hormones, BP, Respiratory | Patients | N=417 | Normalizing the variables | No | XGBoost | | Naturalistic | Classification | Self-report | AUC: 0.89 (XGBoost) |
| **[120]** | 2020 | HRV, PPG, Skin Response | | N=15 | | No | LR, RF, SVM, Other | PLUX BITalino Wearable Sensor | Lab | Classification | Stress-inducing task | Accuracy: 80% (RF) |
| **[114]** | 2018 | HRV, PPG | | | | No | SVM | Polar H7 HRM | Lab | Classification | Self-report | Accuracy: 81% (SVM) |
| **[99]** | 2017 | Questionnaire | Stressed Students | N=249 | | No | NN, RF, Fuzzy, SVM, Other | | Pre-collected data without experiment | Classification | BDI scores | Accuracy: 100% (NN) |
| **[109]** | 2020 | Questionnaire | Depressed and Healthy Dutch citizens | N=11,081 (570 self-reported depression) | | No | XGBoost | | Pre-collected data without experiment | Classification | Self-report | Accuracy: 97.65%, Precision: 95.48%, Recall: 99.87%, F1 score: 0.98 (XGBoost) |
| **[157]** | 2019 | HRV, PPG | | | | No | NN | Firstbeat Bodyguard 2, Smartwatch | Lab | Regression | Heart Rate Monitor (HRM) measurements | RMSE: 28.5 (Regression) |
| **[158]** | 2021 | HR, PPG | | | | No | SVM | PPG Sensor, Arduino UNO | Lab | Classification | Stress-inducing task | Accuracy: 62% (SVM) |
| **[70]** | 2021 | EMG, ECG, HRV | Healthy Drivers | N=16 | | No | SVM | ECG, EMG, Volvo S70 | Lab | Classification | Stress-inducing task | Accuracy: 93.7% (SVM) |
| **[101]** | 2017 | Questionnaire | | | | No | NN, Fuzzy, K-means, SVM | | Pre-collected data without experiment | Classification | Pre-collected dataset for depression | Accuracy: 97% (SVM) |
| **[51]** | 2021 | EEG, ECG, HR | Students | N=24 | Using K-means on EEG energy | No | NN, DT, RF, K-means, KNN, SVM | 19-channel EEG, 12-channel ECG | Lab | Classification | Stress-inducing task | Accuracy: 82% (ANN) |
| **[102]** | 2021 | Questionnaire | Indian people with mental disorder | N=395 | | No | LR, NB, RF, KNN, SVM | Google Forms | Naturalistic | Classification | Self-report | Accuracy: 92.15% |
| **[52]** | 2017 | EEG | Healthy subjects | N=42 | | No | LR, NB, SVM | EEG 128 channels, Electrical Geodesic Net Amps 300 amplifier | Lab | Classification | Stress-inducing task | Accuracy: 94.6% (NB) |
| **[111]** | 2020 | Questionnaire | Pregnant Women | N=508 | | No | RF, SVM | Questionnaire | Naturalistic | Classification | EPDS scores | Accuracy: 80% (SVM) |
| **[42]** | 2020 | HR, EMG, Skin Response, Respiratory, ECG | Drivers | N=17 | | No | LR, DT, LDA/QDA, KNN, SVM | | Lab | Classification | Stress markers in driving task | Accuracy: 75.02% |
| **[107]** | 2018 | Questionnaire | Employees | N=750 | | No | LR, DT, Boosting, RF, KNN | Survey | Naturalistic | Classification | | Accuracy: 75.13% (Boosting) |
| **[98]** | 2020 | Questionnaire | | N=8 | | No | DT, RF | Thermostat, and Passive InfraRed Sensors | Naturalistic | Regression | MOS (SF36) Questionnaire | MSE: 0.17 (RF) |
| **[147]** | 2018 | ECG, HR, Skin Response | | | Multi-task learning with participants as tasks in NN | No | NN, LR, SVM | | Pre-collected data without experiment | Classification | Stress-inducing task | AUROC: 0.91 (NN) |
| **[47]** | 2019 | HRV, EEG, Skin Response, Respiratory | Healthy subjects | N=24 | | No | NN | Polar Wearlink HRM, E243 electrodes | Lab | Classification | Stress-inducing task | Accuracy: 90% (NN) |
| **[121]** | 2019 | ECG, PPG, HRV, Skin Response, Respiratory | Healthy subjects | N=30 | | No | RF, DT, KNN | Shimmer 3 ECG, Empatica E4 | Lab | Classification | Stress-inducing task | Accuracy: 84.13 (RF) |
| **[159]** | 2017 | Acceleration/Body Movement, Skin Response, Other | Patients with MDD | N=12 | | No | LR, Adaboost, RF, Other | Empatica E4, Android Smartphones | Naturalistic | Regression | Clinical score, self-report | RMSE: 2.8 (Ridge) |
| **[86]** | 2020 | HR, HRV, PPG, Questionnaire | Healthy subjects | | | No | LR, DT, RF, SVM | Self-made PPG sensor | Lab | Classification | Audio-visual stress inducing stimulus | Accuracy: 91% (RF), AUROC: 0.96 (RF) |
| **[130]** | 2018 | HR, ECG | | N=15 | | No | SVM, Kalman Filter | | Lab | Classification | Anxiety stimuli | Accuracy: 71% (SVM) |
| **[108]** | 2017 | Questionnaire | Geriatric patients | N=520 | | No | NB, RF, Other | | Lab | Classification | HADS scale | Accuracy: 90%, AUROC: 94.3% (RF) |
| **[142]** | 2021 | Questionnaire | Kenya people | N=800 | | No | LR, NB, DT, Boosting, Adaboost, XGBoost, RF, SVM | | Pre-collected data without experiment | Classification | Survey | Accuracy: 85%, F1 score= 0.78 (SVM, RF, Ada Boosting, Voting Ensemble) |
| **[160]** | 2020 | Questionnaire | People from tech and non-tech companies | | | No | LR, DT, RF, KNN, SVM | Survey | Pre-collected data without experiment | Classification | Survey | Accuracy: 84%, F1 score: 0.87 (LR, DT) |
| **[161]** | 2020 | Questionnaire | Child and Adolescent Twins in Sweden | N=7,638 | | No | NN, LR, XGBoost, RF, SVM | Survey, Reports | Pre-collected data without experiment | Classification | Survey | AUROC: 0.739 (RF) |
| **[162]** | 2019 | Questionnaire | PTSD patients | N=90 | | No | LR, NB, RF, SVM, Ensemble Methods, Hard Voting | Metricwire Mobile app | Naturalistic | Classification | DSM-5 scores | AUROC: 0.85 (Ensemble) |
| **[112]** | 2020 | Questionnaire | Veterans with PTSD | N=305 | | No | LR, Boosting | | Naturalistic | Classification | Self report | F1 score: 0.69 (Voting Classifier) |
| **[163]** | 2021 | Questionnaire | NESDA cohort participants | | | No | LR, NB, AUTO-SKLEARN | | Pre-collected data without experiment | Classification | Self report | Accuracy: 79% (auto-sklearn) |
| **[164]** | 2020 | Questionnaire | Students | N=917 | | No | NN, Boosting, Adaboost, KNN, DT, SVM, RF | WhatsApp | Pre-collected data without experiment | Classification | GAD-7 score | Accuracy: 75.4% (NN) |
| **[105]** | 2021 | Questionnaire | Students | N=4184 | | No | NN, LR, Boosting, XGBoost, RF, KNN, SVM | | Pre-collected data without experiment | Classification | MDD/GAD patients and normal people | AUC: 0.73 GAD, 0.67 MDD (XGBoost) |
| **[141]** | 2021 | ECG | Chinese Academy of Sciences | N=34 | | No | NN | Sticker-Type ECG | Lab | Classification | Stress-inducing task, self-report | Accuracy: 86.8%, Specificity: 0.93 (LSTM) |
| **[165]** | 2019 | HRV, ECG | | | | No | LDA/QDA, SVM | | Lab | Classification | Stress-inducing task | Accuracy: 82.7% (CNN) |
| **[166]** | 2021 | Skin Response | WESAD dataset | N=15 | Only showed the importance of personalization | No | NN | WESAD Dataset | Pre-collected data without experiment | Classification | Stress-inducing task | Accuracy: 92.85% (CNN) |
| **[167]** | 2021 | HRV, ECG, Skin Response | | | | No | DT | | Lab | Classification | Self-report | Accuracy: 79% (SVM) |
| **[82]** | 2020 | Audio Signals, HRV, Skin Response, ECG | WESAD and SWELL datasets | N=50 | One model was trained for each subject (from ground-up and transfer learning) | No | DT | SWELL and WESAD Datasets | Naturalistic | Classification | Questionnaire scores, self-report | Accuracy: 95.2% (RF) |
| **[132]** | 2018 | Skin Response, ECG, PPG, HRV | Employees with reported stress | N=12 | | No | SOM | MindMedia NeXus-10 MKII, imec Health Patch | Lab | Classification | SOMs were used to generates | Accuracy: 79%, Sensitivity: 75.6% |
| **[87]** | 2018 | Text | Depressed people | N~9000 | | No | NN, Boosting, XGBoost | | Pre-collected data without experiment | Classification | Reddit Self-reported Depression Diagnosis (RSDD) dataset | Precision: 16.9%, Recall: 17.8%, F1 score: 17.6% (increased comapred to CNN) |
| **[95]** | 2021 | Questionnaire | | N=39,975 | | No | SVM, RF, XGBoost, DT, NB | DASS Questionnaire | Pre-collected data without experiment | Classification | DASS scores | AUROC: 0.98 (SVM) |
| **[168]** | 2021 | HR | PTSD patients | N=99 | | No | XGBoost, RF, GLM, SVM | Apple Watch, MOTO 360 | Naturalistic | Classification | Self-report | Accuracy: 83%, AUROC: 0.7 (XGBoost) |
| **[59]** | 2017 | Eye Tracking, Acceleration/Body Movement, HR, PPG | Adults | N=23 | | No | Adaboost, KNN, SVM, NB | Camera | Lab | Classification | Stress-inducing task | Accuracy: 88.32% (KNN) |
| **[60]** | 2021 | Eye Tracking | Adults | N=23 | | No | NN (CNN, LSTM) | Camera | Lab | Classification | Stress-inducing task | Accuracy: 86.1% (LSTM) |